\documentclass[english, a4paper, 12pt]{report}
\usepackage{graphicx,amssymb,lineno,url,amsmath,tabularx}
\usepackage{latexsym}
\usepackage{algpseudocode}
\usepackage{amsfonts}
\usepackage{gensymb}
\usepackage{epsfig}

\pdfoutput=1
\DeclareGraphicsExtensions{.pdf,.eps,.jpg,.bmp,.gif,.jpeg}
\usepackage{verbatim}

\usepackage{epstopdf}
\usepackage{listings}
\usepackage{color}
\usepackage{fancyhdr}
\usepackage{float}
\usepackage{setspace}
\usepackage{sidecap}
\definecolor{gray}{rgb}{0.4,0.4,0.4}
\definecolor{darkblue}{rgb}{0.0,0.0,0.6}
\definecolor{cyan}{rgb}{0.0,0.6,0.6}
\usepackage{float}
\usepackage{setspace}
\usepackage{amsmath}
\usepackage{algorithm}

\usepackage{graphicx,geometry, amssymb, lineno, url, lscape, subfigure, , amsfonts, verbatim, latexsym, dsfont, multicol, multirow, array, longtable}

\usepackage{latexsym}
\usepackage{amsfonts}
\usepackage{epsfig}
\usepackage[english]{babel}
\usepackage[table]{xcolor}
\usepackage[normalem]{ulem} % either use this (simple) or
\usepackage{soul} % use this (many fancier options)

\usepackage{cite}
\usepackage{graphicx,amssymb,lineno,url,amsmath,babel}

\usepackage{latexsym}
\usepackage{amsfonts}
\usepackage{epsfig}
\DeclareGraphicsExtensions{.pdf,.eps,.jpg,.bmp,.gif,.jpeg,.pdf}
\usepackage{verbatim}
\usepackage{epstopdf}
\usepackage{listings}

\usepackage{color}
\definecolor{gray}{rgb}{0.4,0.4,0.4}
\definecolor{darkblue}{rgb}{0.0,0.0,0.6}
\definecolor{cyan}{rgb}{0.0,0.6,0.6}

\begin{document}
%	\onehalfspacing

\thispagestyle{empty}

\begin{center}
	
	\vspace{1.5cm}
	
	\textbf{\huge  Automatic Image De-fencing System\\ 
		%\vspace{1.5mm}
	}
	
\end{center}
\vfill
{\flushright \hspace{10cm} { \textit{\large \textbf{N. Krishna Kanth}}}}

\cleardoublepage

\begin{center}
	
	%\vspace{1.5cm}
	
	\textbf{\huge Automatic Image De-fencing System\\ 
		%\vspace{1.5mm}
	}
	\vspace{24pt}
	\center { \textit{\textbf{Report submitted to} }}
	\center{ \textit{\textbf{Indian Institute of Technology, Kharagpur}}}
	\center{ \textit{\textbf{for the award of the degree}}}
	\center{\textit{\textbf{of}}}
	\vspace{12pt}
	
	\textbf{\large Master of Technology }\\
	\vspace{12pt}
	\textbf{\large in Electrical Engineering with Specialization }\\
	\vspace{12pt}
	\textbf{\large in ``Instrumentation \& Signal Processing'' }\\
	
	\vspace{12pt}
	{\large \textit{\textbf{by}}}\\
	
	\vspace{12pt}
	
	\textbf{\large N. Krishna Kanth}\\
	\vspace{12pt}
	\textbf{\large 10EE35022}
	
	\vspace{12pt}

	\begin{figure}[H]
		\begin{center}
			\epsfig{file=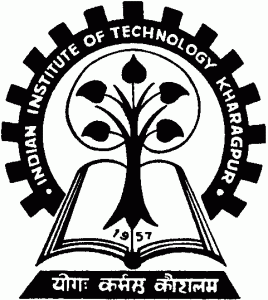, width=30mm}
		\end{center}
	\end{figure}
	\vspace{12pt}
	\textbf{\large DEPARTMENT OF ELECTRICAL ENGINEERING}\\
	\vspace{12pt}
	\textbf{\large INDIAN INSTITUTE OF TECHNOLOGY, KHARAGPUR}\\
	\vspace{12pt}
	\textbf{ May, 2015}\\
	\vspace{0.4cm}
	\copyright 2015, N.Krishna Kanth. All rights reserved. \\
	
\end{center}

\pagenumbering{roman}
\cleardoublepage

\newpage

%\pagestyle{fancy}
%\chapter*{\centering Declaration}
{\bf \Huge \begin{center} {Declaration} \end{center} }
\vspace*{36pt}

\addcontentsline{toc}{chapter}{Declaration}

I certify that
\begin{enumerate}
\item The work contained in this report is original and has been done by me under the guidance of my supervisor

\item  The work has not been submitted to any other Institute for any degree or diploma.

\item I have followed the guidelines provided by the Institute in preparing the
report.

\item I have conformed to the norms and guidelines given in the Ethical Code
of Conduct of the Institute.

\item  Whenever I have used materials (data, theoretical analysis, figures, and text) from other sources, I have given due credit to them by citing them in the text of the report and giving their details in the references. Further, I have taken permission from the copyright owners of the sources, whenever necessary.

\end{enumerate}

\vspace{20mm}  % vertical space

\hspace{80mm}\rule{35mm}{.15mm}\par   % horizontal space, line, start new line
\hspace{80mm} N. Krishna Kanth\newline
\hspace*{90mm} 10EE35022\par

\hspace*{65mm} Department of Electrical Engineering \newline
\hspace*{68mm}Indian Institute of Technology, Kharagpur \newline
\hspace*{85mm}West Bengal, India \newline
Date:\\
Place:
\newpage
\cleardoublepage

{\bf \Huge \begin{center} {Certificate} \end{center} }
\vspace*{36pt}
%\chapter*{\centering Certificate}
\addcontentsline{toc}{chapter}{Certificate}

\begin{figure}[!h]
	\centering
	\includegraphics[height=40mm,width=35mm]{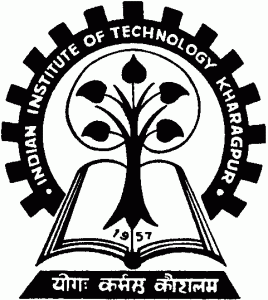}
\end{figure}

\begin{center}

	\vspace{12pt}
	\textbf{\large Department of Electrical Engineering}\\
	\vspace{4mm}
	\textbf{\large Indian Institute Of Technology, Kharagpur}\\
	\vspace{4mm}
	\textbf{\large West Bengal, India}\\

\end{center}
This is certify that the dissertation report entitled, \textbf{\textquotedblleft Automatic Image De-fencing System"} to Indian
Institute of Technology Kharagpur, India, is a record of bonafide project work carried
out by him under my supervision and guidance and is worthy of consideration
for the award of the degree of Master of Technology in Electrical Engineering
with specialization in \textbf{\textquotedblleft Instrumentation and Signal Processing"} of the Institute.
\vspace{20mm}  % vertical space

\hspace{80mm}\rule{48mm}{.15mm}\par   % horizontal space, line, start new line
\hspace{80mm} Prof. Rajiv Ranjan Sahay\par

\hspace*{65mm} Department of Electrical Engineering \newline
\hspace*{68mm}Indian Institute of Technology, Kharagpur \newline
\hspace*{90mm}West Bengal, India \newline
Date:\\
Place:

\cleardoublepage
\newpage

{\bf \Huge \begin{center} {Acknowledgement} \end{center} }
\vspace*{24pt}

\addcontentsline{toc}{chapter}{Acknowledgement}

My heart pulsates with the thrill for tendering gratitude to those persons
who have helped me in workings of the project. No one can even survive, let
alone building a thesis, without countless direct and indirect helps from others.
The most pleasant point of presenting a thesis is the opportunity to thank
those who have contributed to it. Unfortunately, the list of expressions of thank
no matter how extensive is always incomplete and inadequate. Indeed this page
of acknowledgment shall never be able to touch the horizon of generosity of
those who tendered their help to me.
First and foremost, I would like to express my gratitude and indebtedness to
\emph{Prof. Rajiv Ranjan Sahay}, for his inspiring guidance, constructive criticism and
valuable suggestion throughout this project work. I am sincerely thankful to
him for his able guidance and pain taking effort in improving my understanding
of this project.

I would also like to thank \emph{J. Shankar Ganesh}, a research scholar in the department, who closely followed this work. I am highly indebted to him for clarifying my doubts and for providing suggestions. Also, for the help he provided while
collecting fence datasets. Last but not the least, I would like to thank my parents and my friends for their constant support, love and encouragement. Their selfless guidance has helped me to find my path in this beautiful journey called life.

\vspace{20mm}  % vertical space

\hspace{80mm}\rule{35mm}{.15mm}\par   % horizontal space, line, start new line
\hspace{80mm} N. Krishna Kanth\newline
\hspace*{90mm} 10EE35022\par

\hspace*{65mm} Department of Electrical Engineering \newline
\hspace*{68mm}Indian Institute of Technology, Kharagpur \newline
\hspace*{85mm}West Bengal, India \newline
Date:\\
Place:

\cleardoublepage

\vspace*{3.8cm}
{\bf \Huge \begin{center} {Abstract} \end{center} }
\vspace*{24pt}

\addcontentsline{toc}{chapter}{Abstract}

	\thispagestyle{plain}

	Tourists and Wild-life photographers are often hindered in capturing their cherished images/videos by a fence that limits accessibility to the scene of interest. The situation has been exacerbated by growing concerns of security	at public places and a need exists to provide a tool that can be used for post-processing such \textquotedblleft fenced videos” to	produce a \textquotedblleft de-fenced" image. There are several challenges in this problem, we identify them as $1.$ Robust detection of fence/occlusions. 2. Estimating pixel motion of background scenes. 3. Filling in the fence/occlusions by utilizing information in multiple frames of the input video. In this work, we aim to build an automatic post-processing tool that can efficiently rid the input video of occlusion artifacts like fences. Our work is distinguished by two major contributions. The first is the introduction of learning based technique to detect the fences patterns with complicated backgrounds. The second is the formulation of objective function and further minimization through loopy belief propagation to fill-in the fence pixels from multiple frames. We observe that grids of Histogram of oriented gradients descriptor using Support vector machines based classifier and Convolutional based deep neural networks significantly outperforms in terms of  detection accuracy of texels. We present results of experiments using several real-world videos to demonstrate the effectiveness of the proposed fence detection and de-fencing algorithm.

\cleardoublepage

\newpage

\pagestyle{fancy}
%\tableofcontents
\begin{spacing}{0.5} \tableofcontents \end{spacing} 
\newpage
\pagestyle{fancy}
\listoftables
\addcontentsline{toc}{chapter}{List of Tables}
\newpage
\pagestyle{fancy}
\listoffigures

\addcontentsline{toc}{chapter}{List of Figures}

\cleardoublepage

%
%\pagestyle{fancy}
%
%
%%\chapter*{Acknowledgement}
%\addcontentsline{toc}{chapter}{Abstract}
%
%
%\begin{abstract}
%
%	Tourists and Wild-life photographers are often hindered in capturing their cherished images/videos by a fence that limits accessibility to the scene of interest. The situation has been exacerbated by growing concerns of security	at public places and a need exists to provide a tool that can be used for post-processing such \textquotedblleft fenced videos” to	produce a \textquotedblleft de-fenced" image. There are several challenges in this problem, we identify them as $1.$ Robust detection of fence/occlusions. 2. Estimating pixel motion of background scenes. 3. Filling in the fence/occlusions by utilizing information in multiple frames of the input video. In this work, we aim to build an automatic post-processing tool that can efficiently rid the input video of occlusion artifacts like fences. Our work is distinguished by two major contributions. The first is the introduction of learning based technique to detect the fences patterns with complicated backgrounds. The second is the formulation of objective function and further minimization through loopy belief propagation to fill-in the fence pixels. We observe that grids of Histogram of oriented gradients descriptor using Support vector machines based classifier significantly outperforms detection accuracy of texels in a lattice. We present results of experiments using several real-world videos to demonstrate the effectiveness of the proposed fence detection and de-fencing algorithm.
%	
%\end{abstract}

\newpage

\pagenumbering{arabic}
\setcounter{page}{1}
\cleardoublepage
\onehalfspacing

\chapter{Introduction}
\label{sec:intro}
Tourists and amateur photographers capture their cherished moments at historical places or monuments which they visit during their journeys. Travellers are today freed from carrying heavy camera equipment since their phones are light-weight and portable yet capable of advanced photography. The availability of low-cost smartphones/phablets with sophisticated cameras has led to an exponential increase in the number of images or videos captured and shared over the internet. Mobile phones have witnessed great improvements in their operating system software with the popular ones having dedicated \textquoteleft apps\textquoteright \hspace{1pt} which manage image/video acquisition and addition of sophisticated photographic effects after data capture. In a relatively short period, significant improvements in resolution of display and cameras have elevated the quality of videos/images captured. Furthermore, proliferation of internet sites for sharing multimedia content has led to a virtuous cycle further driving the quality of imaging hardware in smartphones.

Despite the advances in technology of such devices, sometimes the amateur photographer is frustrated by unwanted elements in the scene. One such hindrance is the presence of fences or barricades occluding the object which the photographer wishes to capture. In recent times, security concerns have resulted in places of tourist interest being barricaded for protection. Fences have become common, restricting access to the public affecting the aesthetic experience of the tourist who wants to preserve his memories for posterity using images/videos.  Sometimes fences are necessary to protect the spectator from grave danger such as wild animals in zoos. Yet one would prefer to enhance the aesthetic appeal of the captured images of these animals by removing  interfering fences/barricades. Therefore, there exists a need for a post-processing tool for seamless removal of occlusions in such images.

The problem of image \textquotedblleft de-fencing" is basically, removal of fences/barricades from an image affected by such occlusions. Recently, there has been considerable progress in the area of image inpainting \cite{Bertalmio,Bugeau,Criminisi,Getreuer,James,Fadili,Wexler,Zongben,Yang,Ruzic,Patwardhan,Komodakis} in which most works assume that the pixels to be filled-in are known and marked by the user in the input image. It is to be noted that for the problem at hand, we cannot make such an assumption since the number of pixels that belong to the fence are too many and it is very tedious to mark them by hand. Hence, a robust automatic technique for detecting the fence in the frames of the input video is required. 
	
In this work, we propose a supervised learning based approach to detect fences/occlusions and an optimization framework to obtain a de-fenced image using a few frames  from the video of the occluded scene. The basic idea is to capture a short video clip of the scene by panning the camera and use a few frames from it to restore data hidden by the fence in the reference image. It is natural to assume that pixels occluded in the reference image are uncovered in the additional frames of the video which is captured by panning the landscape. We observe that although the problem appears simple, it becomes challenging when the scene is dynamic and three-dimensional in nature. 

\begin{figure}[h]
	
	\centering
	\subfigure[Original frame from video]{\includegraphics[width=5.5cm]{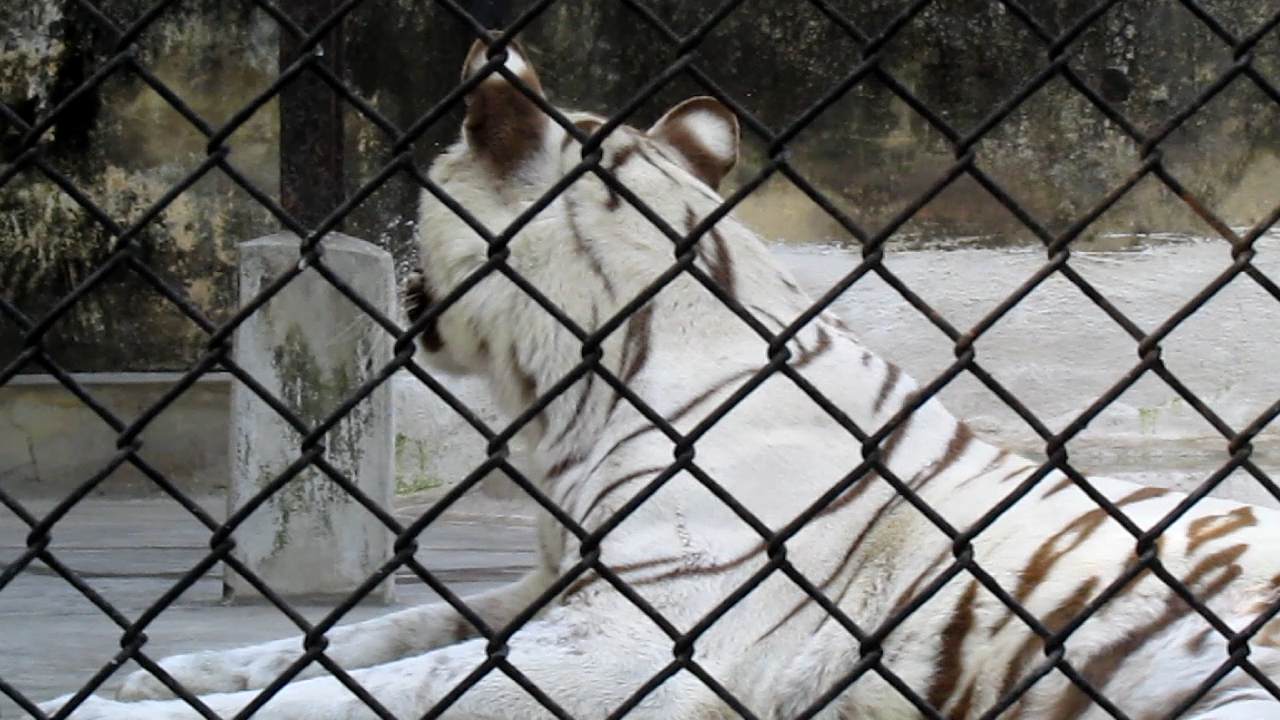}} \hspace{1cm}
    \subfigure[another frame from video]{\includegraphics[width=5.5cm]{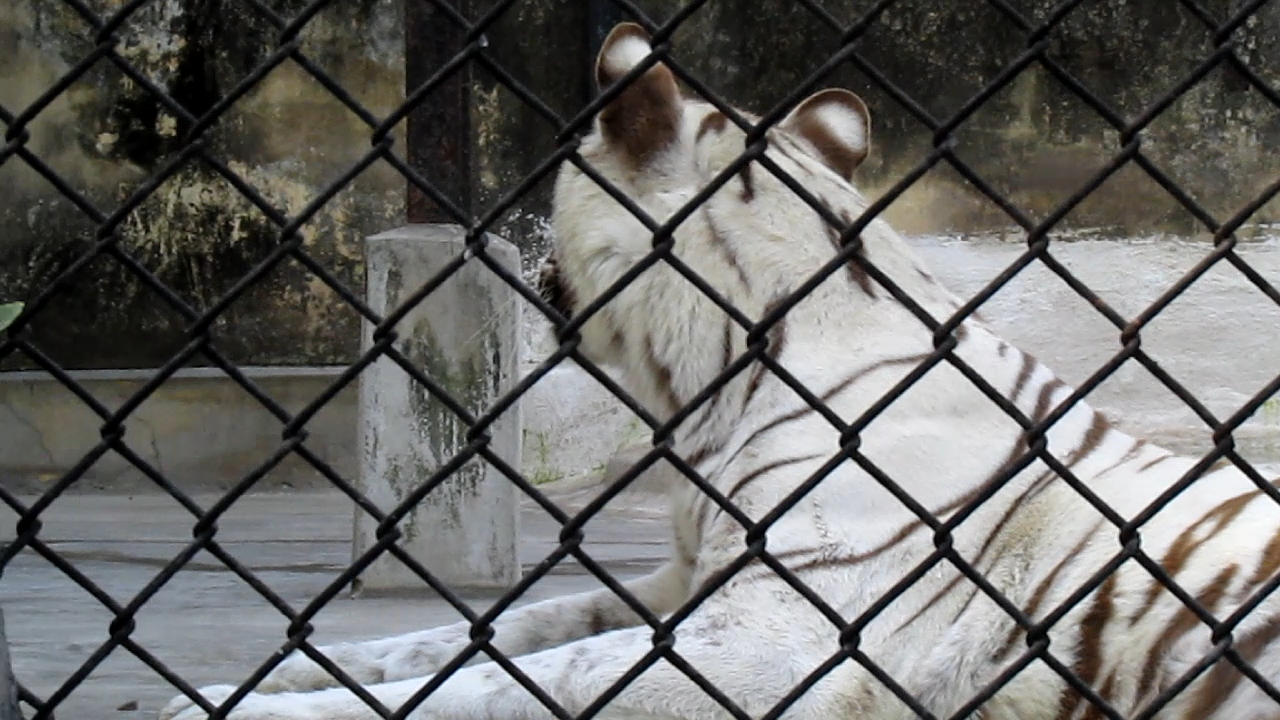}}\\
	\subfigure[Fence detection]{\includegraphics[width=5.5cm]{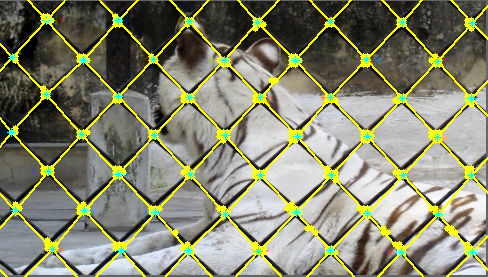}} \hspace{1cm}
	\subfigure[De-fenced image]{\includegraphics[width=5.5cm]{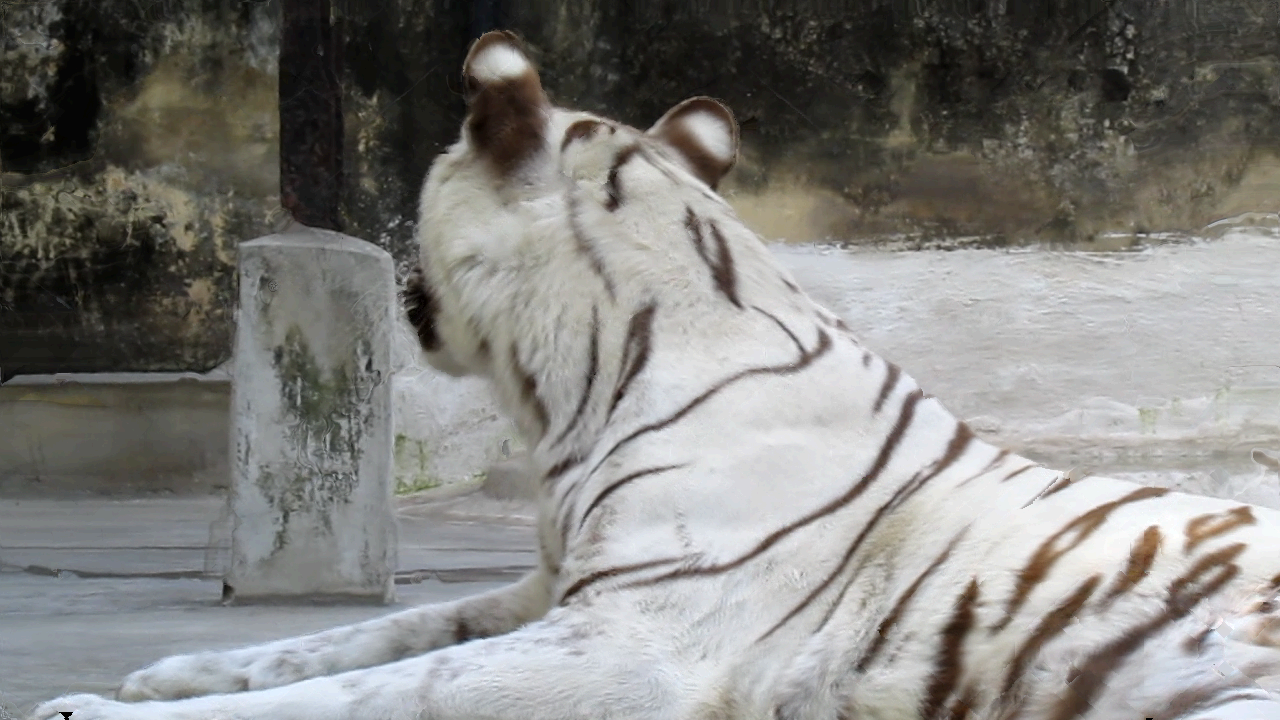}}
	
	\caption{ Lion in a zoo: De-fenced image restored from original fenced video}
	\label{fig:introlion}
\end{figure}

In Figs. \ref{fig:introlion} (a) and (b), two frames from a captured video is shown wherein the fence is occluding parts of the lion's body.  We observe that the motion cue in video can be exploited to perform  \textquotedblleft de-fencing" of the degraded frames to obtain an image wherein the fence has been removed. In Fig.~\ref{fig:introlion} (c), we show a sample output of the proposed algorithm which has successfully removed the occlusions due to fence pixels.

 Unfortunately, there is no benchmark fence data set available in the literature for evaluating the proposed learning based algorithm. For demonstrating the effectiveness of the proposed learning based algorithm we used two different data sets. Firstly, we have collected a dataset consisting of $200$ real-world images/videos under diverse scenarios and complex backgrounds by using a mobile camera (Google Nexus 5). Secondly, we used a subset of images from Penn State University (PSU) near-regular texture (NRT) database \cite{Minwoo}.

Initially, we assumed that the frames of the input video obtained by panning the occluded scene are shifted globally. Hence, we used the affine scale-invariant feature transform (ASIFT) \cite{Guoshen} image descriptor to match corresponding points in the frames obtained from the captured video. In order to overcome the global motion limitation, we use a recently proposed technique \cite{Brox} for motion estimation in videos of dynamic scenes.  The final step in our method is fusion of data from additional frames to \textquotedblleft de-fence" the reference image. For this purpose, we use a degradation model to describe the formation of images affected by occlusions due to fences by  modeling image as a Markov random field (MRF). The task of image de-fencing is posed as an inverse problem. Our approach is more accurate than mere image inpainting since we use data from neighboring frames to derive the maximum a-posteriori (MAP) estimate of the \textquotedblleft de-fenced" image. 
	
The report is organized as follows. We review methods for inpainting, fence detection and de-fencing in Section 2. In Section 3, we outline details of proposed methodology of our algorithm. In Section 5, we discuss the various method of fence detection using machine learning approach and Section 6 explains about the optimization framework for data fusion. Experimental results and comparisons with the state-of-the-art algorithms are given in Section 7 and finally, conclusions are given in Section 8.

\chapter{Literature survey}

\section{Fence detection}
There has been significant research in the field of regular or near-regular pattern detection \cite{yanxi_pami2004,Hays,Wen-Chieh,Minwoo,Minwo}. Initially, the authors in \cite{yanxi_pami2004} formulated the computational model for periodic pattern perception based on the theory of frieze and wallpaper groups. An effective fence-detection algorithm is proposed in \cite{Minwoo}.Recently, there has been significant progress in addressing the problem of detection of near-regular textures in images\cite{Park}. However, the method in \cite{Park} is not able to robustly detect all fence pixels. Traditional texture filling tools such as Criminisi et al. \cite{Criminisi} require users to manually mask out unwanted image regions, but this process would be tedious (taking hours) and error-prone. Simple color-based segmentations are not sufficient. Interactive foreground selection tools such as Lazy Snapping are also  not well suited to identifying the thin, web-like structures. 

The authors in \cite{Minwoo} detect regular textures in three phases. Firstly, in phase 1, the template lattice is discovered. In the second phase, mean shift belief propagation is used (MSBP) to spatially scan the image and expand the lattice. In the third phase, the deformed lattice is rectified into a regular lattice using regularized thin-plate spline warping. Later, the work of \cite{James} posed lattice discovery as a higher order correspondence problem and discovered patterns with significant texel variations. Recently, Yadong et al. \cite{Yadong} addressed the video de-fencing problem wherein they proposed soft fence detection method by using visual parallax as the cue to distinguish fences from the un-occluded pixels. This cue is reasonable upto some extent where the scenes consist of static elements. However, this algorithm failed to detect fences in videos of real-world dynamic scenes.   

\section{Image inpainting/de-fencing}

Image inpainting is one of the challenging problems in image processing since many years. In the literature, image inpainting techniques are classified into two broad categories, namely diffusion-based methods and exemplar-based techniques. Diffusion-based techniques \cite{Bertalmio,Ballester,Levin,Bertalmio_cvpr,Roth} use smoothness priors to propagate information from known regions to the unknown region. These algorithms work satisfactorily if the region with missing data is small in size and low-textured in nature. However, it is difficult to fill large occluded regions and recreate fine texture using diffusion-based techniques. On the contrary, exemplar-based techniques \cite{James,Criminisi_cvpr,Criminisi,Zongben} fill-in the occluded regions by using similar patches from other locations in the image. Methods belonging to this category possess the advantage that they can recreate texture missing in large regions of the image.

Many authors addressed the inpainting problem either using structure propagation or by using texture synthesis. But Criminisi et al. \cite{Criminisi} combined the advantages of both approaches and proposed an algorithm for filling-in large regions. The work of Kedar et al. \cite{Patwardhan} addressed video inpainting under constrained camera motion. The authors of \cite{Papafitsoros} approached the image inpainting problem using a variational framework, wherein an energy function has been formulated with a data fidelity term and regularization prior. The regularizer used in \cite{Papafitsoros} is the combination of both $L1$ as well as $L2$ norms of the inpainted image. The inpainted image is obtainted as the minimization of the proposed  energy function using split Bregman iterations. Kaimeng et al.\cite{Kaiming} added another novel aspect to exemplar-based inpainting techniques wherein they used the statistics of patch offsets. Recently, Tijana et al. \cite{Ruzic} proposed a context aware inpainting algorithm by using normalized histogram of Gabor responses as the contextual descriptors. They have formulated an MRF-based optimization framework in \cite{Ruzic} for the inpainting problem and the same has been solved using an inference algorithm. 

Liu et al. in \cite{Yanxi_cvpr} first addressed the de-fencing problem via inpainting of the occluded foreground pixels. In their method, the fence mask is detected using a regularity dicovery algorithm proposed in \cite{Hays}. The filling-in of the occluded pixels is attempted by using the algorithm of \cite{Criminisi}. Basically, \cite{Yanxi_cvpr} treats the de-fencing problem as an inpainting problem. Subsequently, the authors in \cite{Minwo} extended image de-fencing using mutliple images, which significantly improves the performance due to availability of hidden information in additional frames. The work in \cite{Minwo} proposed a learning based algorithm to improve the accuracy of lattice detection and segmentation. The algorithm collects the positive samples at the patches centered at lattice points and negative sample from in-between locations. Using both the samples, RGB color histograms were computed at each sample location and finally represented as features. They trained an optimal classifier using the features obtained from corresponding foreground masks. Finally, the classification with minimum error is concluded to the best one and corresponding foreground mask is selected which is further used in inpainting.

The authors in \cite{Khasare} proposed an improved multi-frame de-fencing algorithm by using loopy belief propagation. However, there are two issues with that approach. Firstly, the work in \cite{Khasare} assumed that motion between the frames is \textit{global}. This assumption is invalid for more complex dynamic scenes. Also, the method used an image matting technique proposed by \cite{Yuanjie} for fence detection which involves \textit{significant user interaction}. Here, in this work we addressed the above issues by exploiting the nature of fences and addressed fence detection using supervised learning. In order to overcome the assumption of global motion we also considered many real-word videos from YouTube to validate our algorithm wherein the motion is non-global.

\chapter{Proposed methodology} 

Our approach for image de-fencing necessitates the solution of three sub-problems, which we identified as
\begin{enumerate}
	\item  Automatic detection of spatial locations of fences or occlusions in the frames of the video
	\item  Accurate estimation of relative motion between the frames
	\item  Data fusion to fill-in occluded pixels in the reference image with uncovered scene data in additional frames 
\end{enumerate}

In this work we have proposed a novel and robust technique for fence detection. This method utilizes supervised learning to infer pixels belonging to occlusions/fences in the scene. After the fences/occlusions have been identified, we need to fill-in the missing information in order to \textquotedblleft de-fence" the reference frame. A naive idea is to simply inpaint the fenced reference image by a standard image completion technique. However, such an approach would approximate missing information by propagating neighboring pixel intensities respecting edges/discontinuities in the frame. Importantly, by resorting to in-painting techniques we do not exploit the important fact that relative motion between the camera and the scene can cause additional frames to contain data that is missing in one frame. On the other hand, in order to exploit the availability of additional data, we need to estimate motion between the frames accurately.

	\begin{figure}[t]
		\centering
		{\includegraphics[width=13cm]{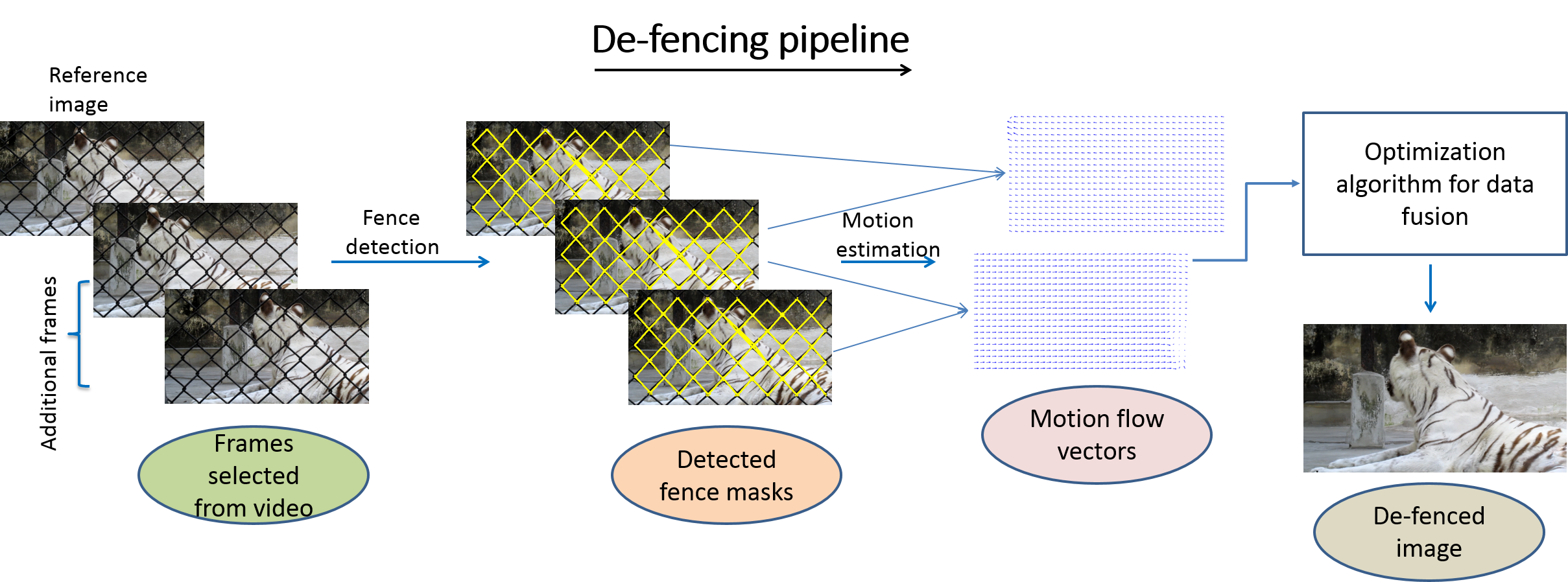}}
		\caption{Image de-fencing pipeline}
		\label{fig:workflow}
		
	\end{figure}

Initially, we assumed that the frames of the input video obtained by panning the occluded scene are shifted globally. Hence, we used the affine scale-invariant feature transform  image descriptor to match corresponding points in the frames obtained from the captured video. By avoiding false matches with a little user interaction, it is possible to estimate the pixel motion accurately. The above assumption restricts our algorithm to videos having global motion only and so it is not general enough for real-world data wherein scene elements could also be dynamic. In order to overcome this limitation we need an algorithm which estimates local pixel motion automatically. Therefore, we use a recently proposed optical flow \cite{Brox} technique for motion estimation in videos of dynamic scenes.

The final step in our method is fusion of data from additional frames to de-fence the reference image. For this purpose, we use a degradation model to describe the formation of images affected by occlusions due to fences. The de-fenced image is modeled as a Markov random field. The task of image de-fencing is posed as an inverse problem.We use the loopy belief propagation technique to optimize an appropriately formulated objective problem. Our approach is more accurate than mere image inpainting since we use data from neighboring frames to derive the maximum a-posteriori estimate of the de-fenced image.proposed algorithm is shown in Fig. \ref{fig:workflow} and the detailed analysis is discussed in the following chapters 

\chapter{Fence detection}

\begin{figure}[h]
	
	\centering
	\subfigure[]{\includegraphics[width=3cm]{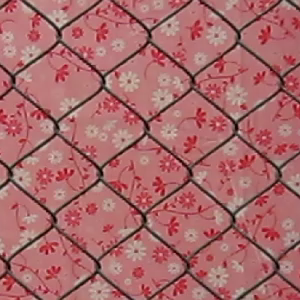}} \hspace{1cm}
	\subfigure[]{\includegraphics[width=3cm]{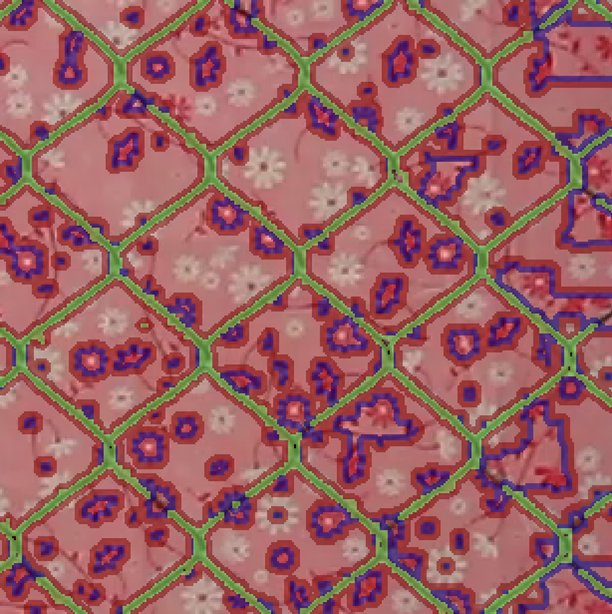}}\\
	
	\label{fig:graphcuts}
	\caption{ Pin paper (a) is the frame from video (b) GraphCut segmentation result}
\end{figure}

In the early stages, we solved the problem of fence detection using two methods. First and simplest way is to treat the fence detection as a segmentation problem. We employ the graph-cuts image segmentation algorithm proposed by \cite{Kolmogorov} on Fig.~\ref{fig:graphcuts}. This approach works in some cases, however for real world problems robust segmentation algorithms fail if the foreground and background layers are of similar color. Secondly, if one considers the fence as the foreground then we can leverage on the process made in the area of image matting through a learning-based approach. The algorithm works with user interaction and used to extract the foreground pixels.

Recently, there has been significant progress in addressing the problem of detection of near-regular textures in images \cite{Park}. We observe that fences/barricades can be classified as near-regular textures and used the algorithm proposed in . However, the method in \cite{Park} is not able to robustly detect all fence pixels. To automate our de-fencing system, we proposed an automatic technique for fence detection using a learning based approach.

\begin{figure}[h]
	
	\centering
	\subfigure[]{\includegraphics[width=3cm]{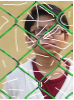}} \hspace{1cm}
	\subfigure[]{\includegraphics[width=3cm]{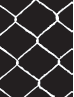}}\\

	\caption{(a) Scribbles put by the user  as
		input for the algorithm in \cite{Yuanjie}. (b) Binary mask denoting the
		fence pixels.}
	\label{fig:vru}
\end{figure}

Also, if one considers the fence as the foreground then we can leverage on the progress made in the area of image matting \cite{Levin}. Recently, a learning-based approach has been proposed for image matting \cite{Yuanjie}. The performance of this algorithm \cite{Yuanjie} is much better and hence, we also used it to completely specify the foreground (fence) pixels. We show the scribbles put by the user on the observation, Fig.~\ref{fig:vru} (a). This scribbled observation is fed as input to the technique in \cite{Yuanjie} whose output is a gray-scale intensity image representing the alpha matte. We threshold this image to obtain a binary mask which denotes the locations of fence pixels as shown in Fig.~\ref{fig:vru} (b). It is to be noted that this scheme for detecting the fence is quite general and does not depend upon the shape/pattern of the particular fence/occlusions. However this approach requires lot of manual interaction and might take lot of time if the fence pattern is dense

\section{Near regular texture segmentation}

Park et al. \cite{Park} has developed an algorithm to detect near regular textures like fences. The procedure is divided into two phases, where the first phase proposes one (t1,t2)-vector pair and one texture element, or texel. It was proved in 2D lattice theory tells us that every 2D repeating pattern can then be reconstructed by translating this texel along the t1 and t2 directions. During phase one, KLT corner features were detected, and the largest group of similar features were selected in terms of normalized correlation similarity. Then the most consistent (t1,t2)-vector pair is proposed through an iterative process of randomly selecting 3 points to form a (t1,t2) pivot for RANSAC and searching for the pivot with the maximum number of inliers.

At phase two, tracking of each lattice point takes place under a 2D Markov Random Field formulation with compatibility functions built from the proposed (t1,t2)-vector pair and texel. The lattice grows outwards from the initial texel locations using the (t1,t2)-vector pair to detect additional lattice points. The tracking is initiated by predicting lattice points using the proposed (t1,t2)-vector pair under the MRF formulation. The inferred locations are further examined; if the image likelihood at a location is high, then that location becomes part of the lattice. However, for robustness, the method avoids setting a hard threshold and uses the region of dominance idea. This was particularly important since there is no prior information about how many points to expect in any given image. We compare our proposed approach over the state-of-the-art algorithm \cite{Park} in following sections.

\section{Machine learning based approach}

\begin{figure}[h]
		\centering
		{\includegraphics[width=13cm]{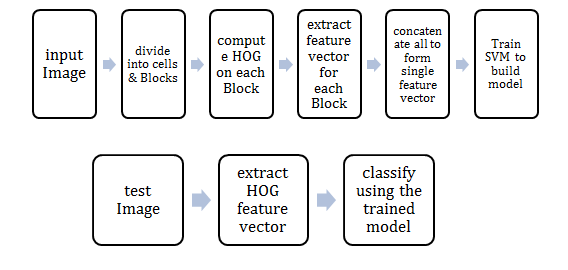}}

		\caption{ Overall  workflow of training and testing stages for machine learning}
		\label{fig:featurehog}
	\end{figure}

In early stages, the fence texels are detected by building a classifier using the RGB features \cite{Minwo}. It was experimentally found that those features are not good enough to classify accurately in difficult cases. In a real-world scenarios, fence texels are rhombic and have joints at their center. We rely on this cue to propose a learning based approach to detect joint positions and later connect them. It is amply demonstrated in the literature that Histogram of Oriented Gradient(HOG) \cite{Dalal} features have been successful in many detection and object classification problems. The HOG feature extraction process and the overall flow of training and testing is shown in Fig.~\ref{fig:featurehog}.  The Fence Detection is done in  two stages. The first stage is training phase to build a classifier model and the second stage is testing phase to predict the output label of the test image.

\subsection{Training phase}

The major steps in involved in the training process are discussed below:

\textbf{1. Concept of cell, block and window}

\begin{figure}[h]
	
	\centering
	\subfigure[fence joints]{\includegraphics[width=5.5cm]{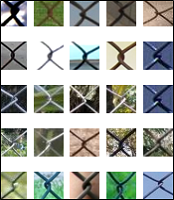}} \hspace{1cm}
	\subfigure[Non-joints]{\includegraphics[width=5.5cm]{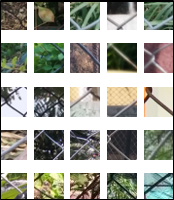}}\\

	\caption{Dataset images (a positives (b) negatives}
	\label{fig:data}
\end{figure}

Every image in the training set is divided into cells of size $4\times4$ pixels. A region of $4$ cells is clustered together to form a single block of size $16\times16$ pixels. Two neighboring blocks has an overlap of $50\%$ of $2$ cells. Hence, a training image of size $30\times30$ pixels is divided into $36$ blocks. We used our own database which consists of $2000$ positive and $6000$ negatives all in the same size of $30\times30$ pixels. The positive and negative dataset sample images are shown in Figs.~\ref{fig:data} (a), (b).

 \textbf{2. Gradient computation}

For every cell in the training image, the gradient magnitude and orientation were computed in both x and y directions. Sobel Filter of mask $3\times3$  is used to calculate the gradients. Before calculating the gradients, the given input image is converted into gray-scale  and Gaussian smoothing is performed  on every cell.  Histogram normalization is done as a preprocessing step to make the training invariant to illuminations. Experiments have proved that these preprocessing steps have little effect has detection rate improvement. 

 \textbf{3. Spatial and orientation binning}
 
Each pixel orientation is discretized into either one of the $9$ Histogram bins. The orientation bins are evenly spaced over $0\degree-180\degree$ with each bin of size $20\degree$. Based on orientation, only one of its bin images is filled with gradient magnitude and rest of them are assigned zero and use them to compute efficiently the HOG. For example, if a pixel at position $(2,2)$ has gradient magnitude $1.2$ and  orientation  $50\degree$, then we fill  pixel $(2,2)$ of Bin $3$  with same magnitude $1.2$  and for rest of the bins are assigned value $0$ at that position. We then compute the integral image for each bin which allows very fast evaluation of HOG features for any rectangular region. The feature vectors from all the blocks are concatenated to form a single feature vector of size $1296$.

 \textbf{4. Training the Support vector Machines}

\begin{figure}[h]
 	\centering
 	{\includegraphics[width=7cm]{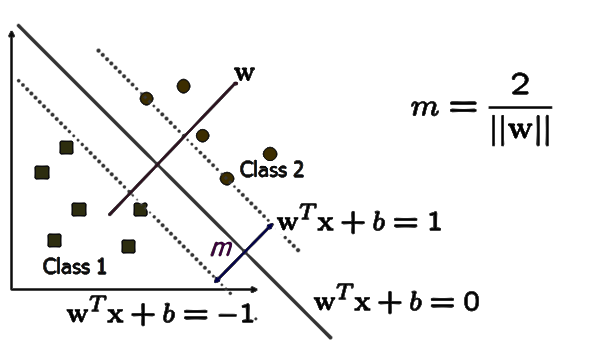}}

 	\caption{Optimal hyperplane with maximum margin}
 	\label{fig:svmp}
 \end{figure}

 The main function of SVM is to map all features into high dimensional space and draw an hyperplane that would best classify the dataset. SVM is a supervised learning machine learning algorithm. The decision boundary / hyperplane should be as far away from the data of both classes as shown in Fig.~\ref{fig:svmp}. Open CV library has  inbuilt SVM based on the implementation of LIBSVM. SVM draws support vectors in high dimensional space  with a weight associated in it and these support vectors are used to predict while testing. SVM Parameters used for training SVM type (CSVC) and  kernel type (RBF) with a 5 fold cross validation set. The parameter C and gamma in RBF kernel is found by iterating on a logarithmic grid and selected based error rate estimated on a 5-fold cross validation. The termination criteria depends on the given number of iterations to be performed and also the maximum tolerance. 
 
 		\begin{table}[H]

 			\centering
 			\begin{tabular}{c c c c }
 				\includegraphics[width=3.5cm,height =2.5cm]{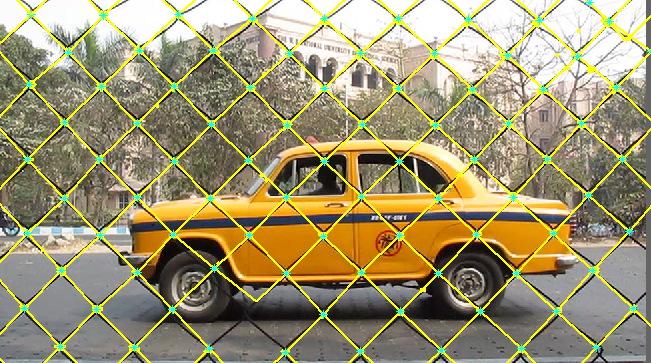} &
 				\includegraphics[width=3.5cm,height =2.5cm]{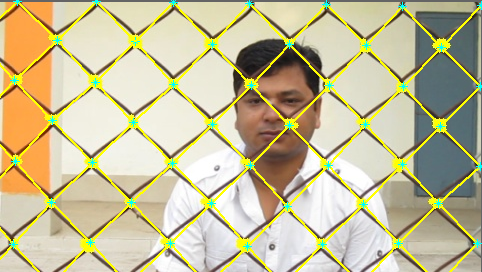} &
 				\includegraphics[width=3.5cm,height =2.5cm]{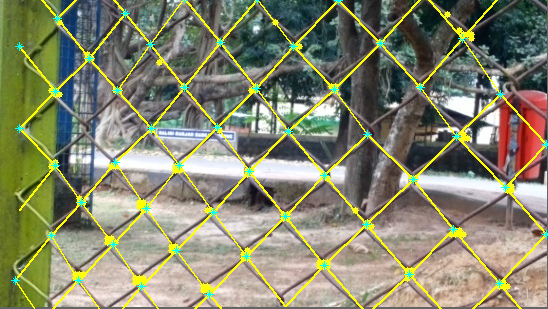} &
 				\includegraphics[width=3.5cm,height =2.5cm]{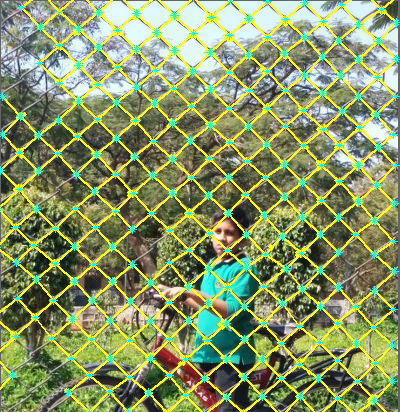} 		\\
 				\includegraphics[width=3.5cm,height =2.5cm]{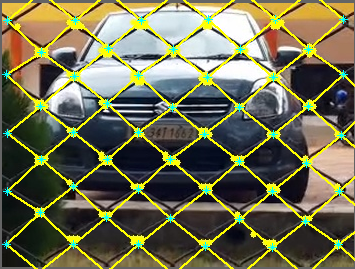} &
 				\includegraphics[width=3.5cm,height =2.5cm]{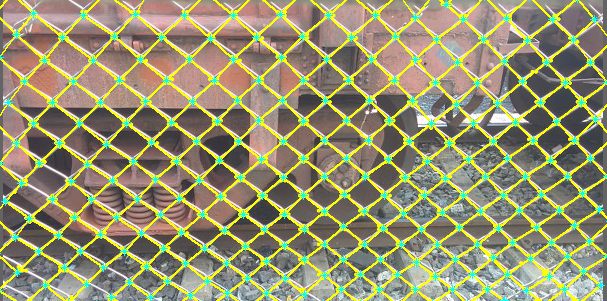} &
 				\includegraphics[width=3.5cm,height =2.5cm]{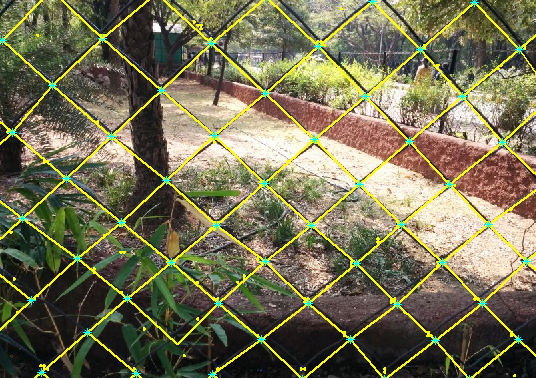} &
 				\includegraphics[width=3.5cm,height=2.5cm]{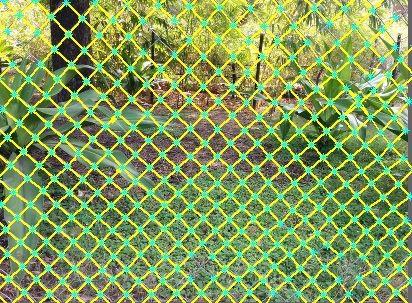} 		\\
 				\includegraphics[width=3.5cm,height =2.5cm]{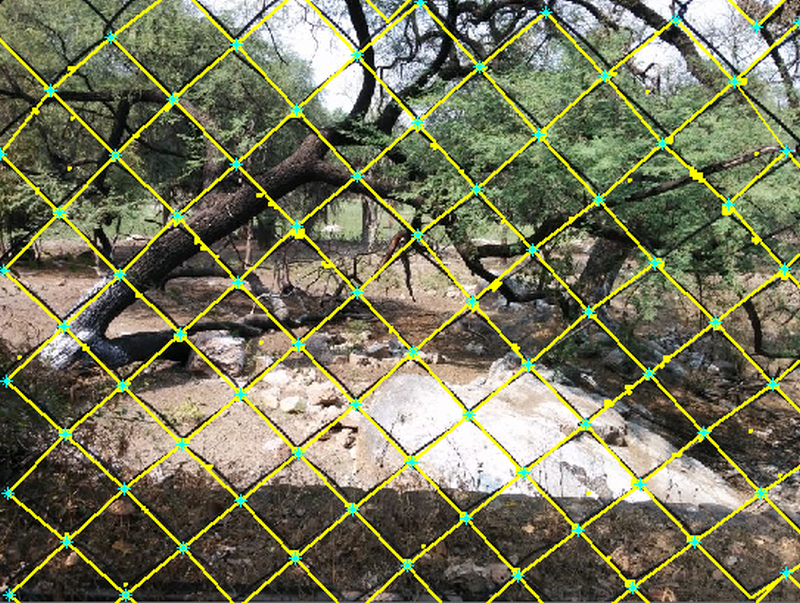} &
 				\includegraphics[width=3.5cm,height =2.5cm]{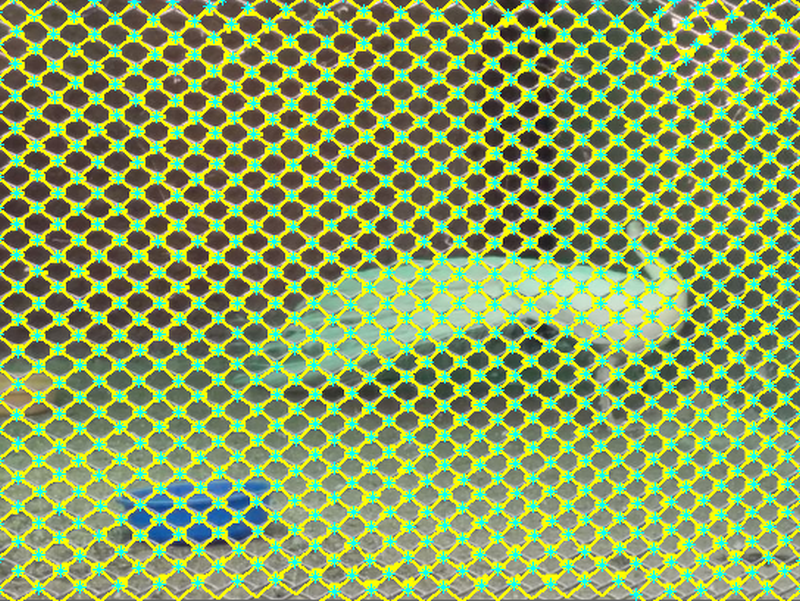} &
 				\includegraphics[width=3.5cm,height =2.5cm]{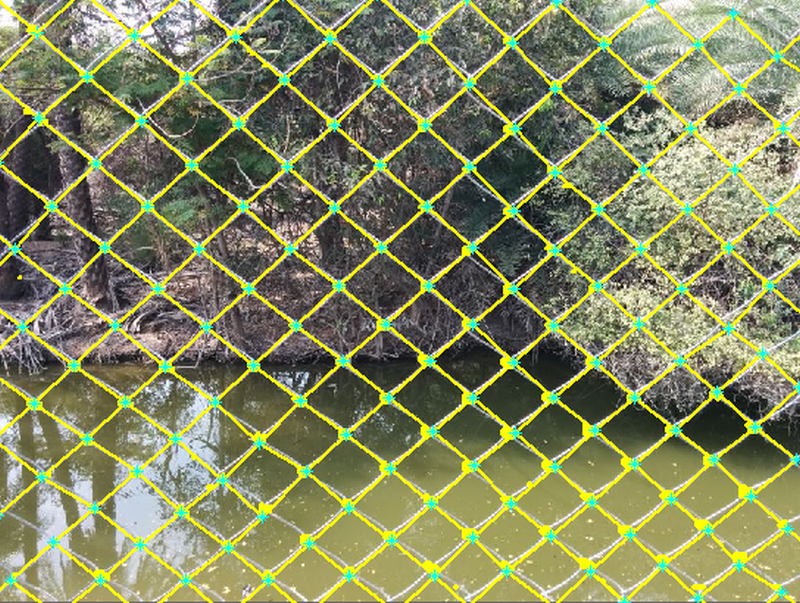} &
 				\includegraphics[width=3.5cm,height =2.5cm]{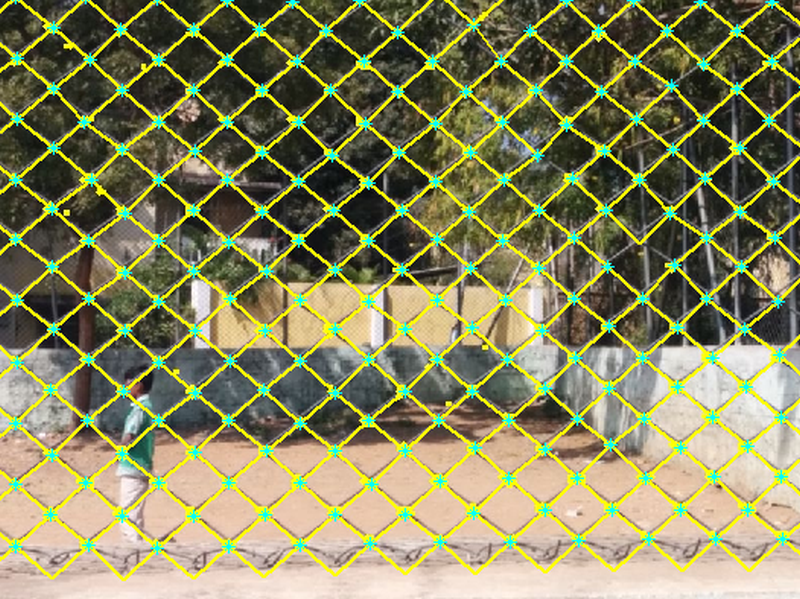} \\
 				\includegraphics[width=3.5cm,height =2.5cm]{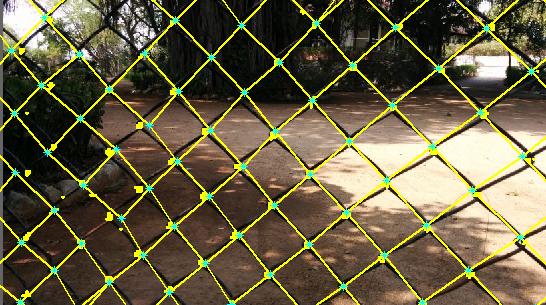} &
 				\includegraphics[width=3.5cm,height =2.5cm]{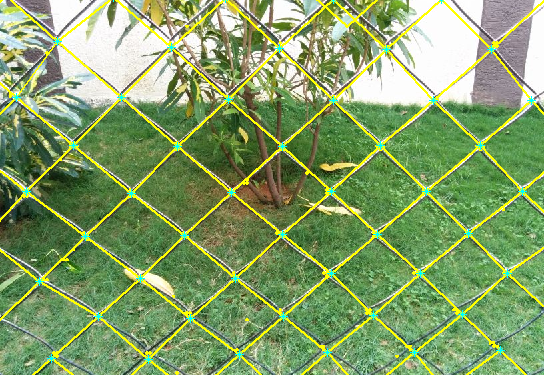} &
 				\includegraphics[width=3.5cm,height =2.5cm]{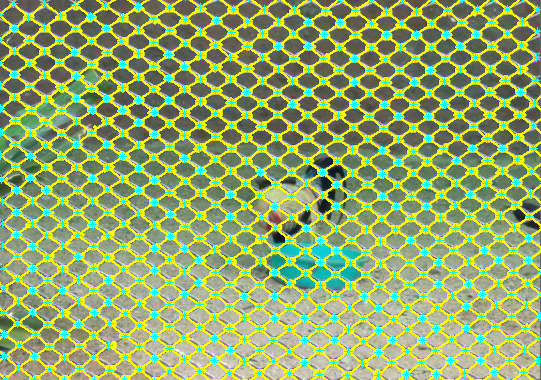} &
 				\includegraphics[width=3.5cm,height =2.5cm]{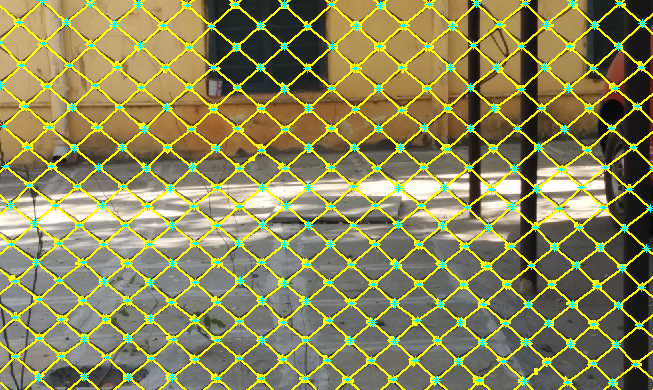} \\
 				\includegraphics[width=3.5cm,height =2.5cm]{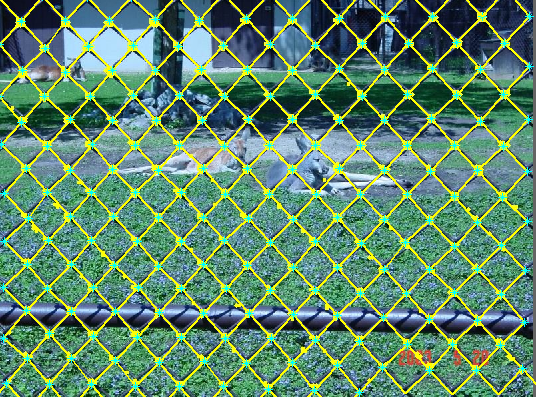} &
 				\includegraphics[width=3.5cm,height =2.5cm]{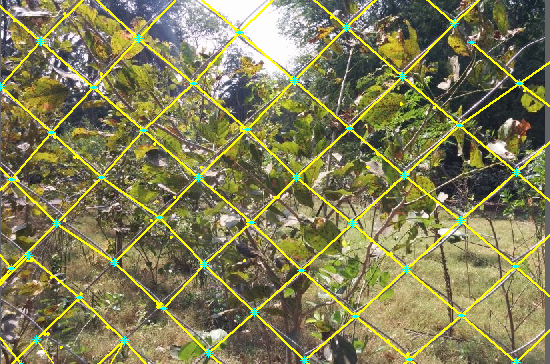} &
 				\includegraphics[width=3.5cm,height =2.5cm]{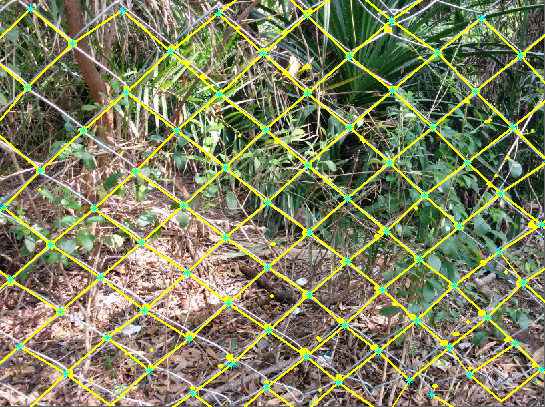} &
 				\includegraphics[width=3.5cm,height =2.5cm]{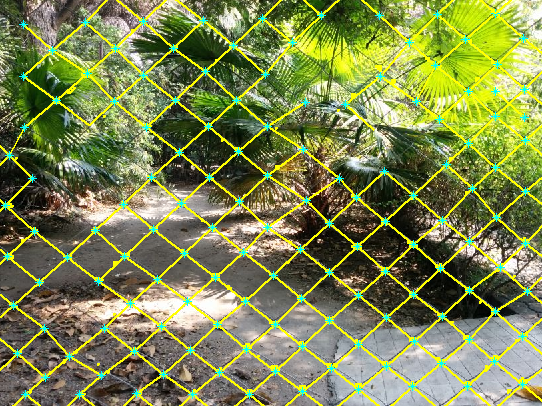} \\
 				\includegraphics[width=3.5cm,height =2.5cm]{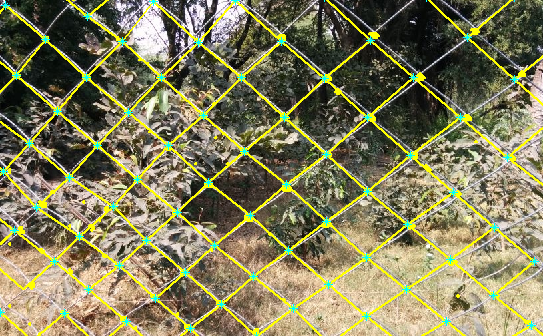} &
 				\includegraphics[width=3.5cm,height =2.5cm]{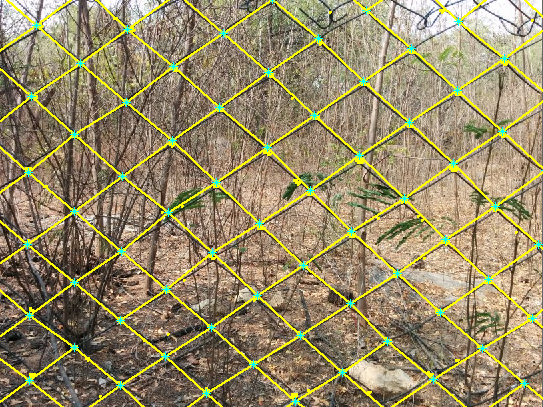} &
 				\includegraphics[width=3.5cm,height =2.5cm]{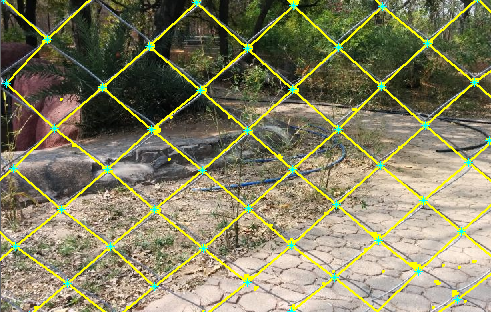} &
 				\includegraphics[width=3.5cm,height =2.5cm]{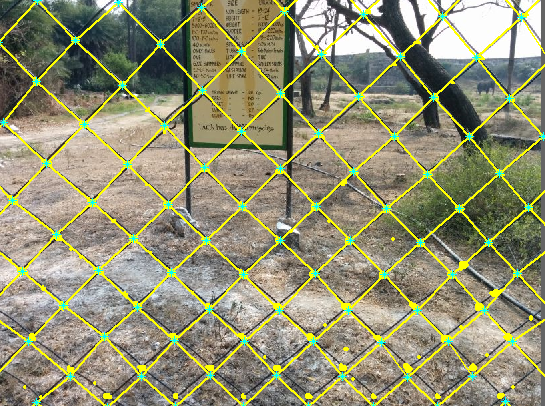} \\
 				
 			\end{tabular}
 			\caption{Detected fences using machine learning approach}
 			\label{tab:fenceresults}
 			
 		\end{table}

 \subsection{Testing phase}
 During detection phase, we scan the test image in sliding window fashion from top to bottom and left to right at different scales with scale ratios of $1.2$ between two consecutive scales. For each detector window, the HOG features are extracted and SVM model is used to classify the joint. Now to connect those joints, we calculate inter joint distances along horizontal and vertical directions. The median of those inter-joint distances is considered as dimension of individual texel. For every detected joint, we find the positive joints within a region of half of inter-joint distance with a reasonable threshold around it. The false positives are from the detected joints are automatically eliminated as those will be located at outside the considered search space. Finally we connect the valid true positive joints to obtain final fence mask and the results of fence detection on data collected from various scenarios are shown in  Table~\ref{tab:fenceresults}.

 \begin{table}[H]

 	\centering
 	\begin{tabular}{c c c}
 		\includegraphics[width=4cm,height =3cm]{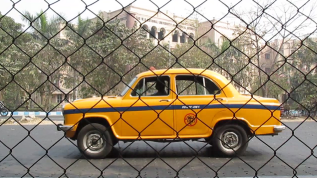} &
 		
 		\includegraphics[width=4cm,height =3cm]{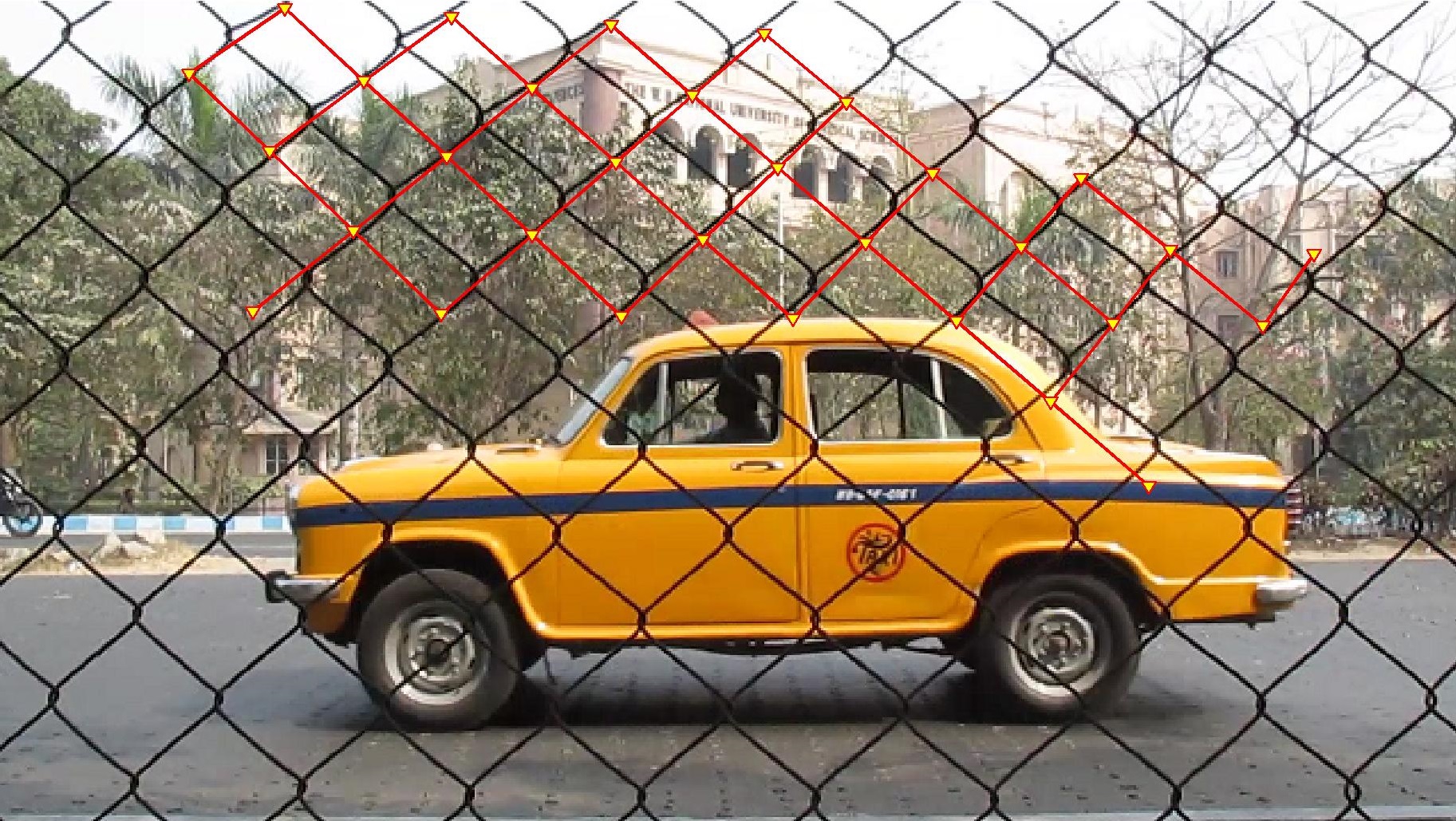}&
 		\includegraphics[width=4cm,height =3cm]{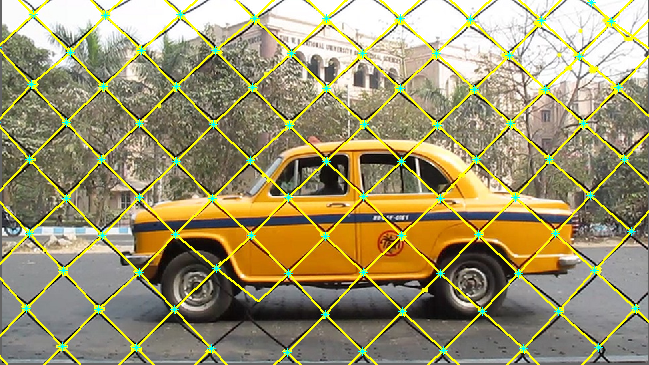} \\
 		
 			\includegraphics[width=4cm,height =3cm]{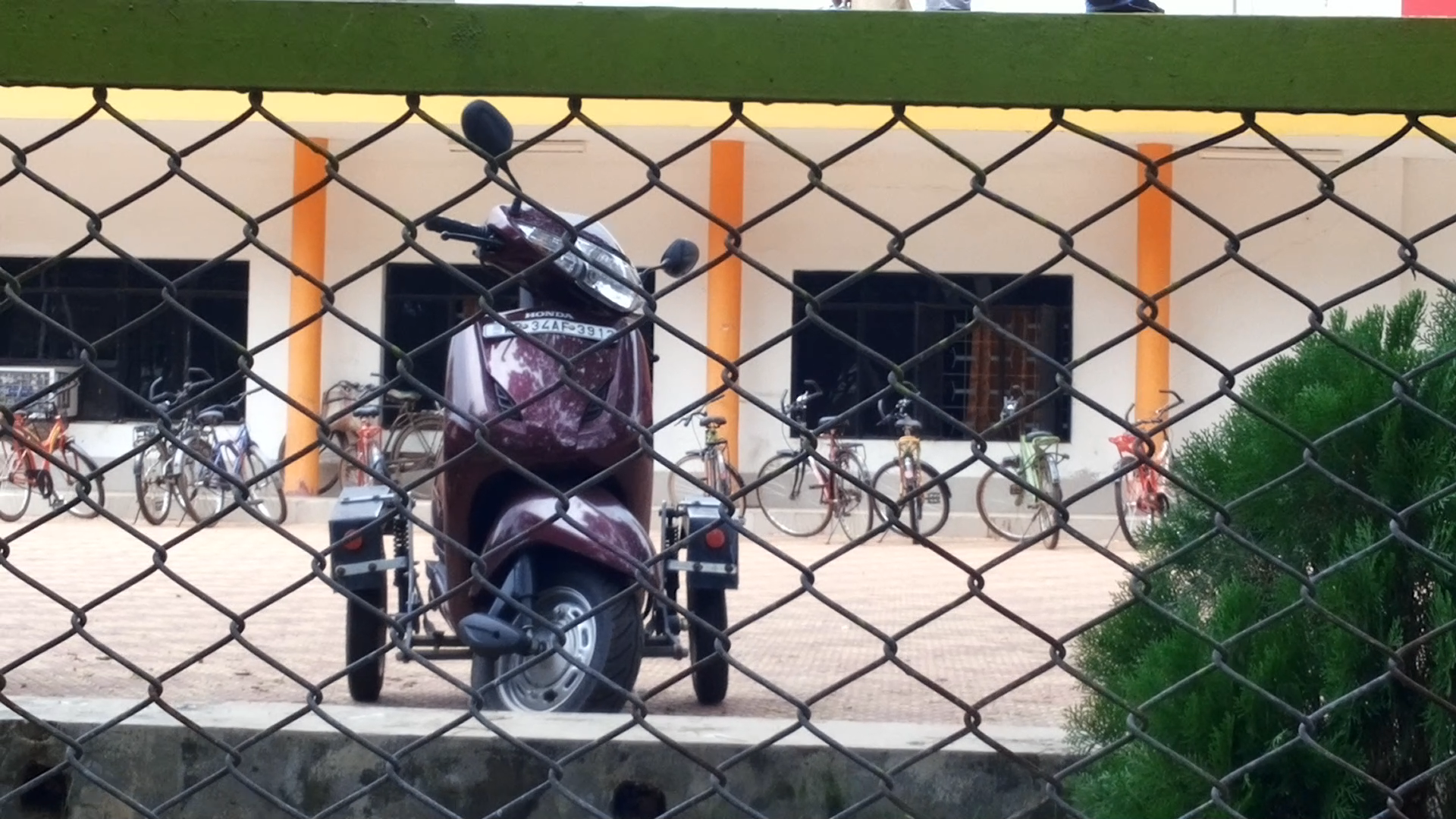} &
 			\includegraphics[width=4cm,height =3cm]{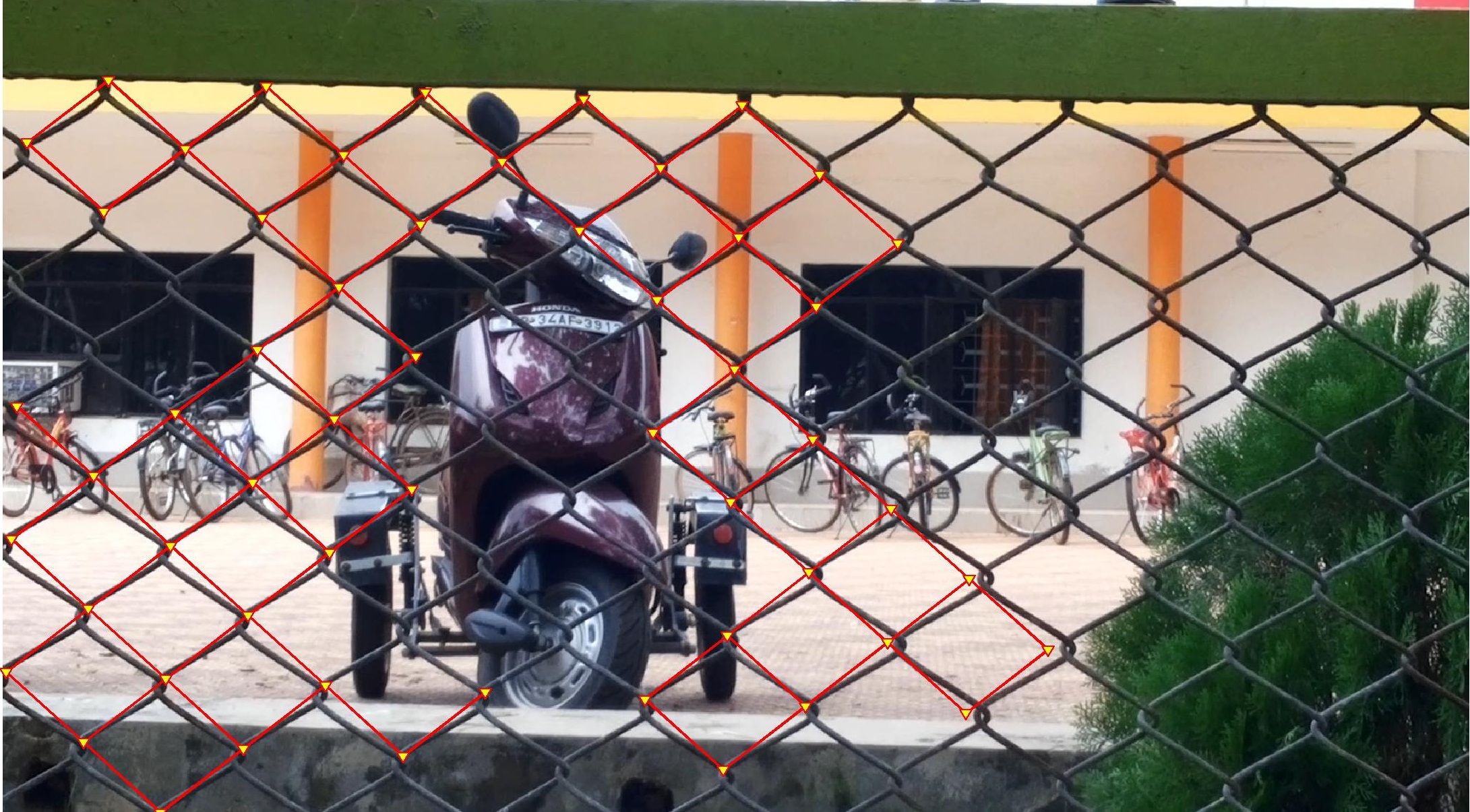} &
 			\includegraphics[width=4cm,height =3cm]{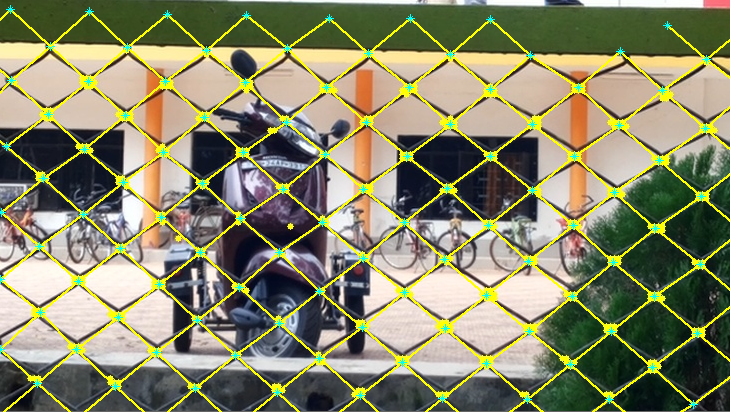} \\
 			
 			\includegraphics[width=4cm,height =3cm]{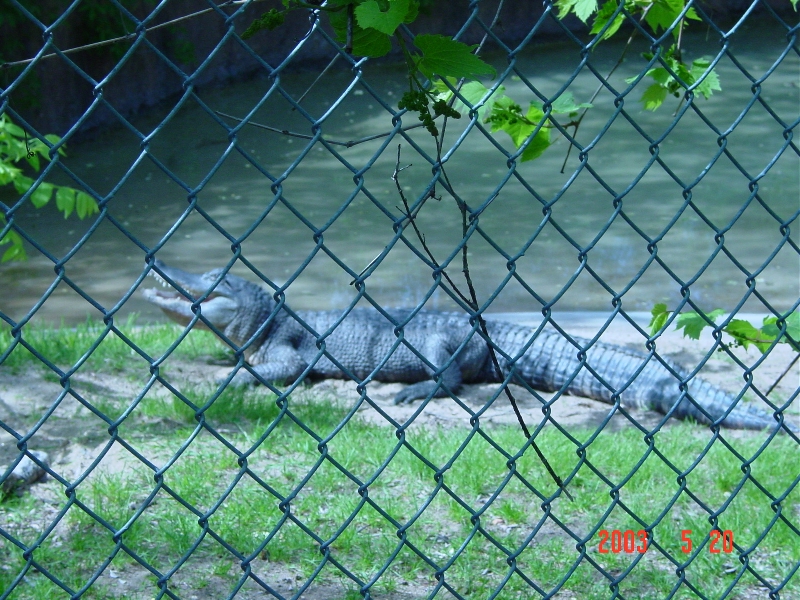} &
 			\includegraphics[width=4cm,height =3cm]{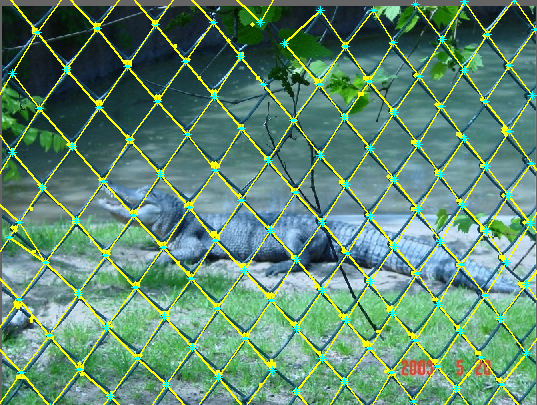} &
 			\includegraphics[width=4cm,height =3cm]{d3.png} \\
 			
 				\includegraphics[width=4cm,height =3cm]{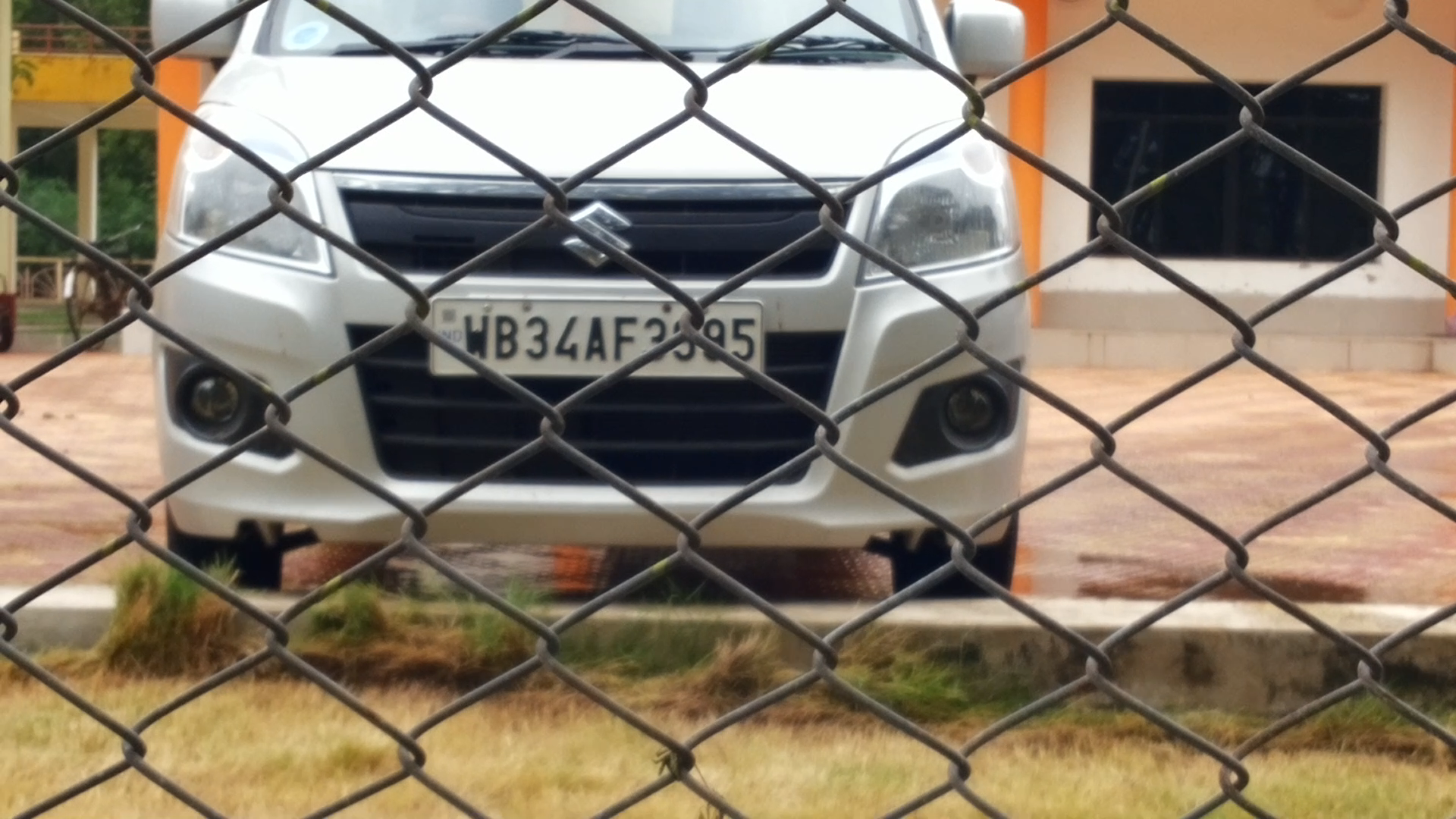} &
 				\includegraphics[width=4cm,height =3cm]{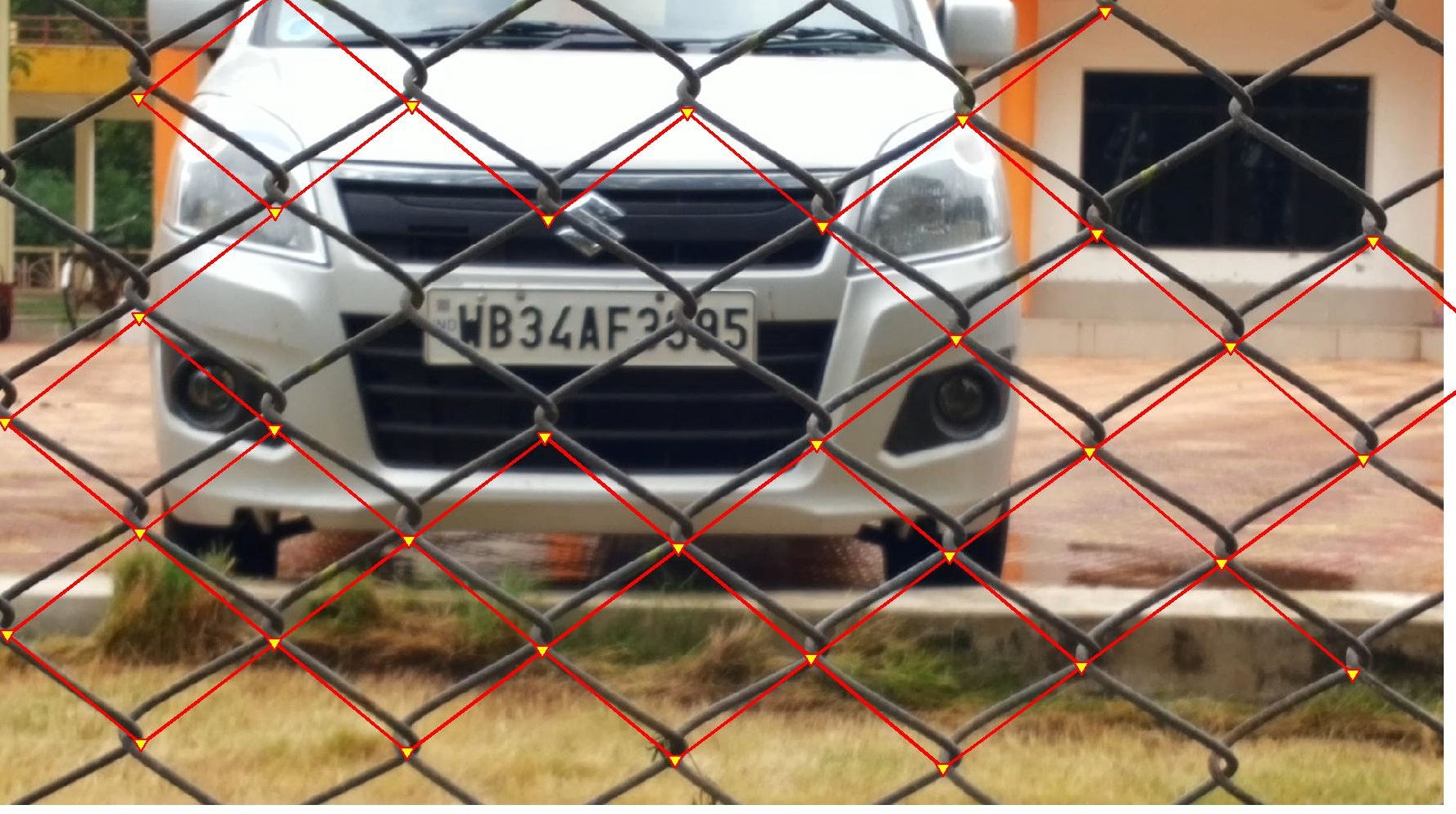} &
 				\includegraphics[width=4cm,height =3cm]{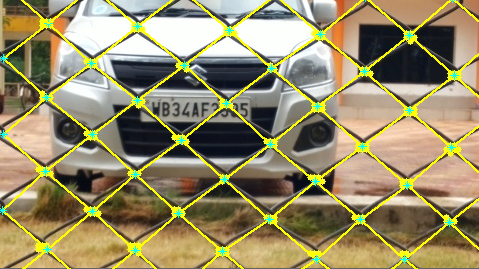} \\
 			\includegraphics[width=4cm,height =3cm]{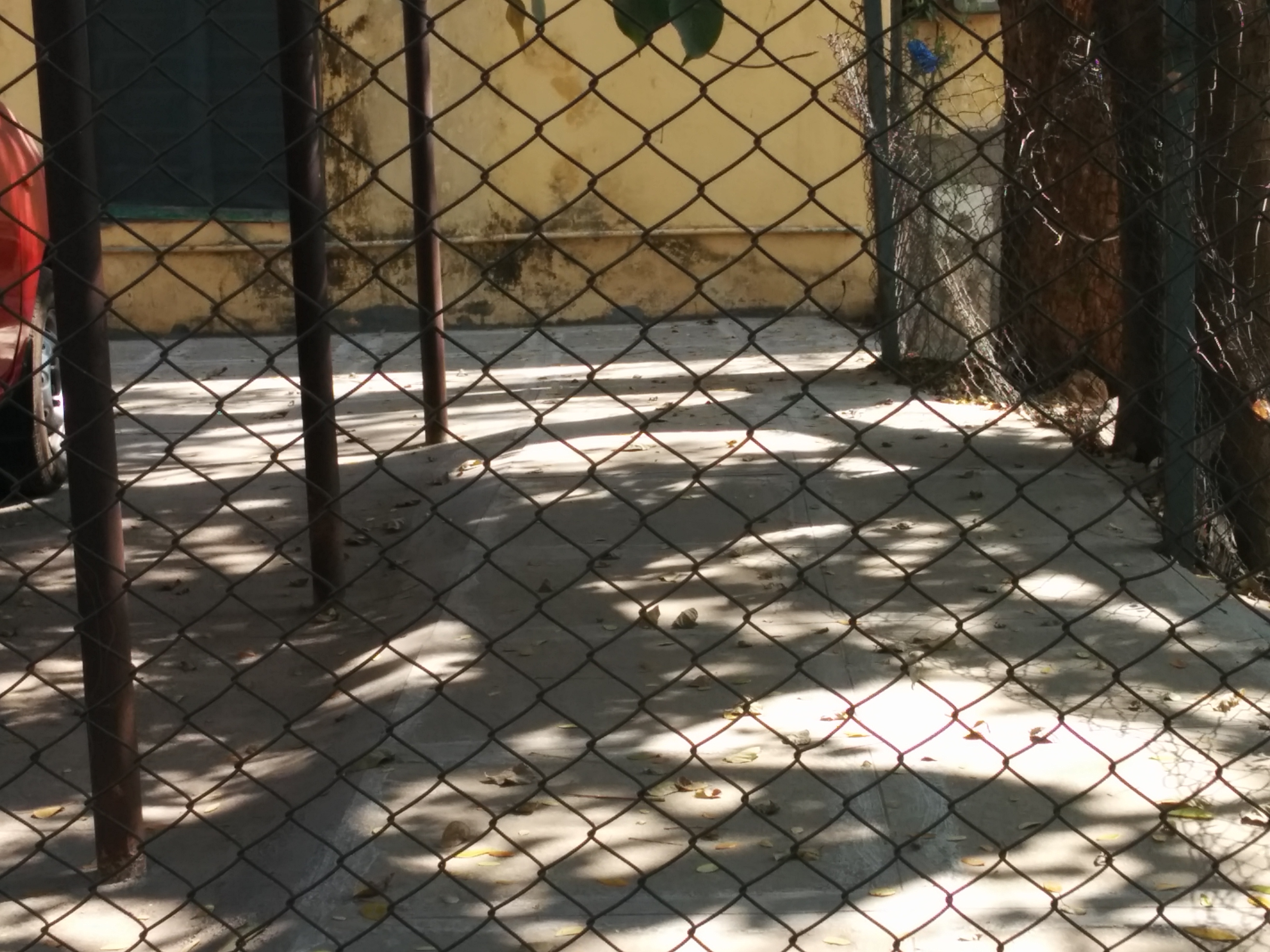} &
 		\includegraphics[width=4cm,height =3cm]{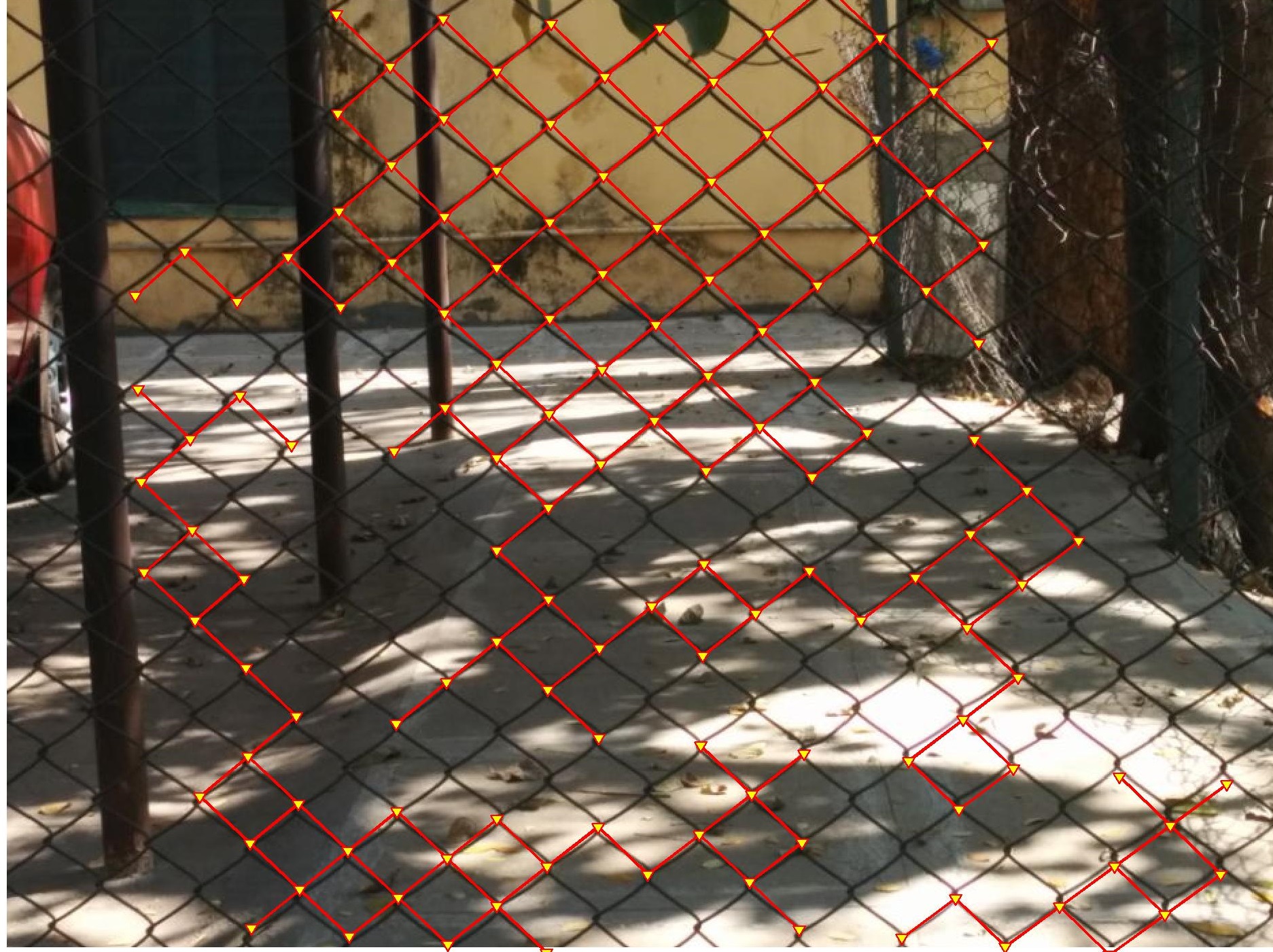} &
    \includegraphics[width=4cm,height=3cm]{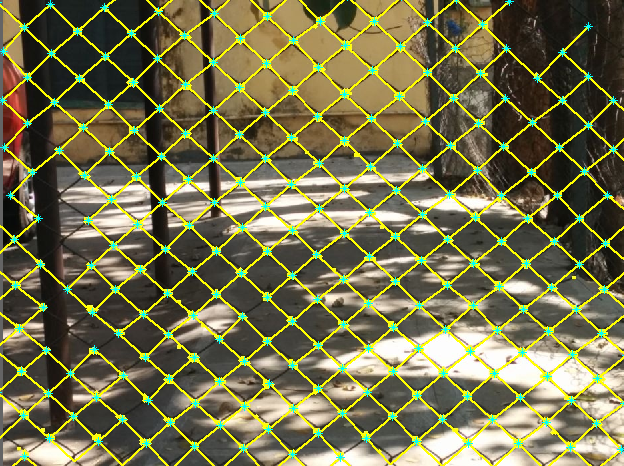} \\	
    
    	\includegraphics[width=4cm,height =3cm]{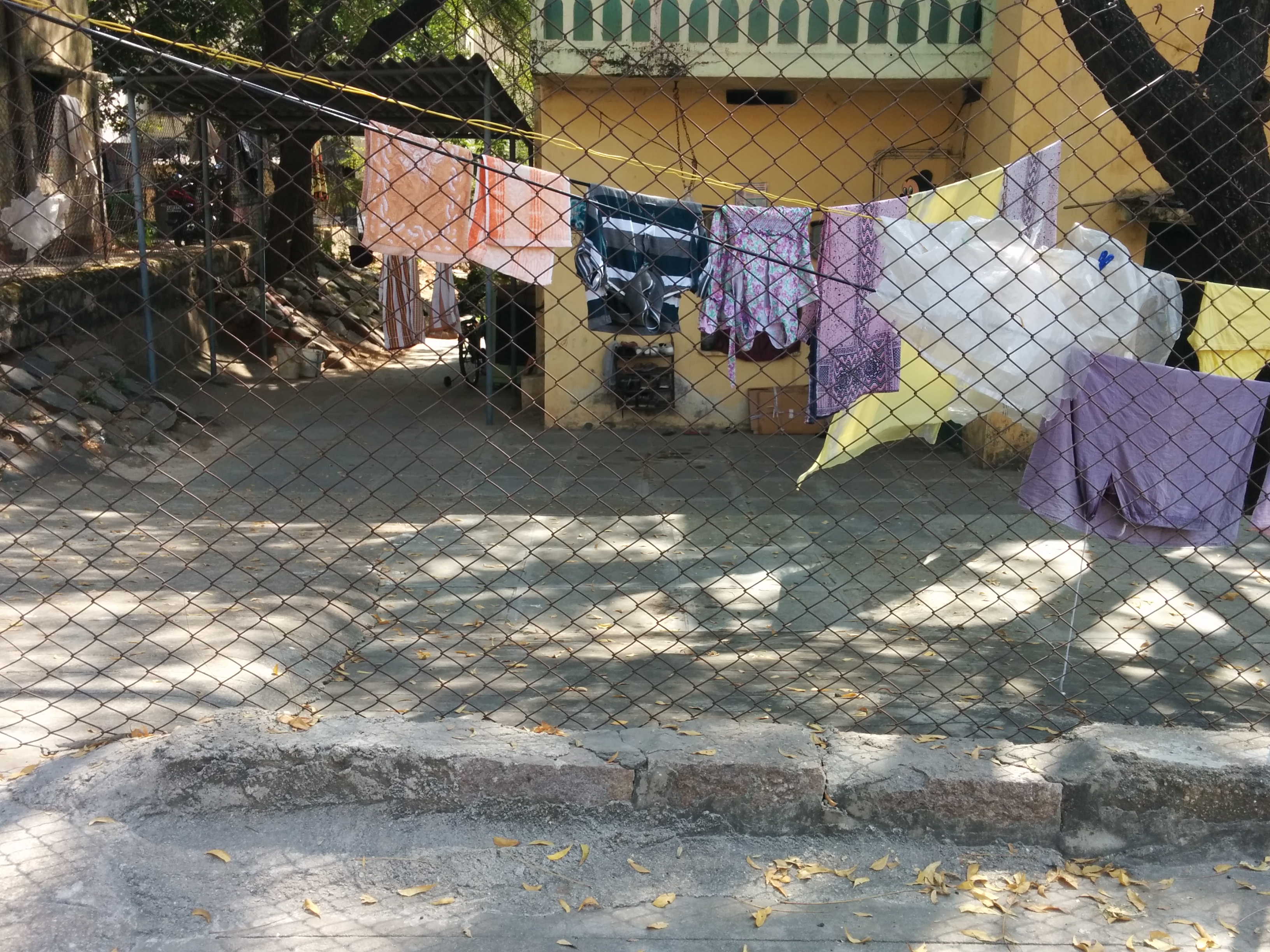} &
    	\includegraphics[width=4cm,height =3cm]{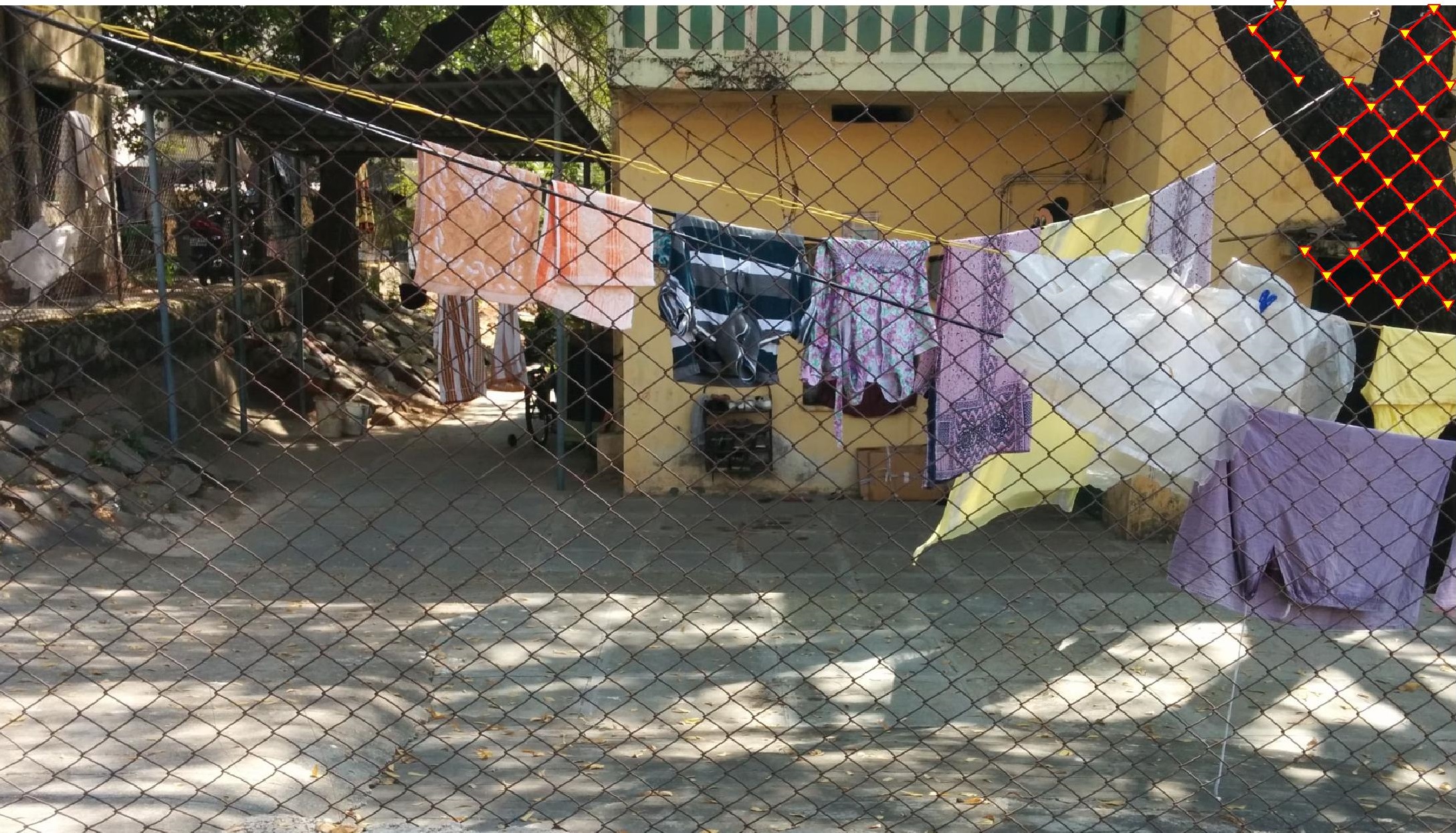} &
    \includegraphics[width=4cm,height=3cm]{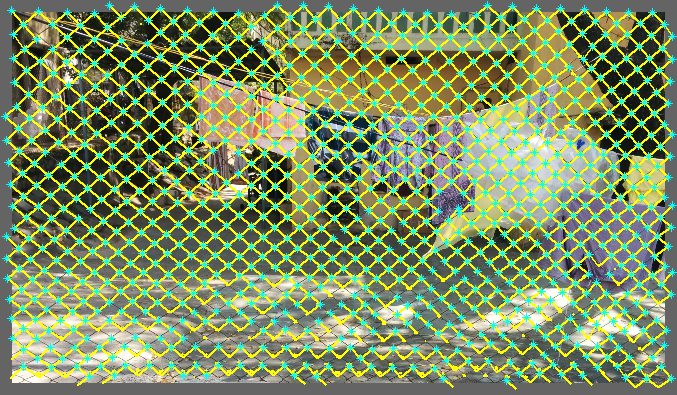} \\

 	\end{tabular}
 	\caption{ Comparison results: column 1 represents original images, 
 	column 2 represents the results of \cite{Park} and column 3 shows the results of our approach	 }
 	\label{tab:fenceresults}
 	
 \end{table}

\textbf{Comparison with state-of-the-art}		
 
 We compare the results of fence detection with \cite{Park} on various databases and report in Table~\ref{tab:fenceresults}. It can be observed that our method significantly performs better in most of the real world scenarios and other technique \cite{Park} failed to propagate the fence patterns on the whole image.

\section{Deep learning}

In the recent studies, it is found that Convolutional neural networks based architecture significantly outperforms the state-of-the-art algorithms in classification problems. The latest generation of CNN have achieved impressive results in challenging benchmarks on image recognition and object detection, significantly raising the interest of the community in these methods. It has been extensively proved the several useful properties of CNN-based representations, including the fact that the dimensionality of the CNN output layer can be reduced significantly without having an adverse effect on performance. In particular, we also performed various data augmentation techniques on base dataset and observed significant performance boost. 

	\subsection{Architecture}
	
		\begin{figure}[h]
			\hfill
			{\includegraphics[height=5cm,width=16cm]{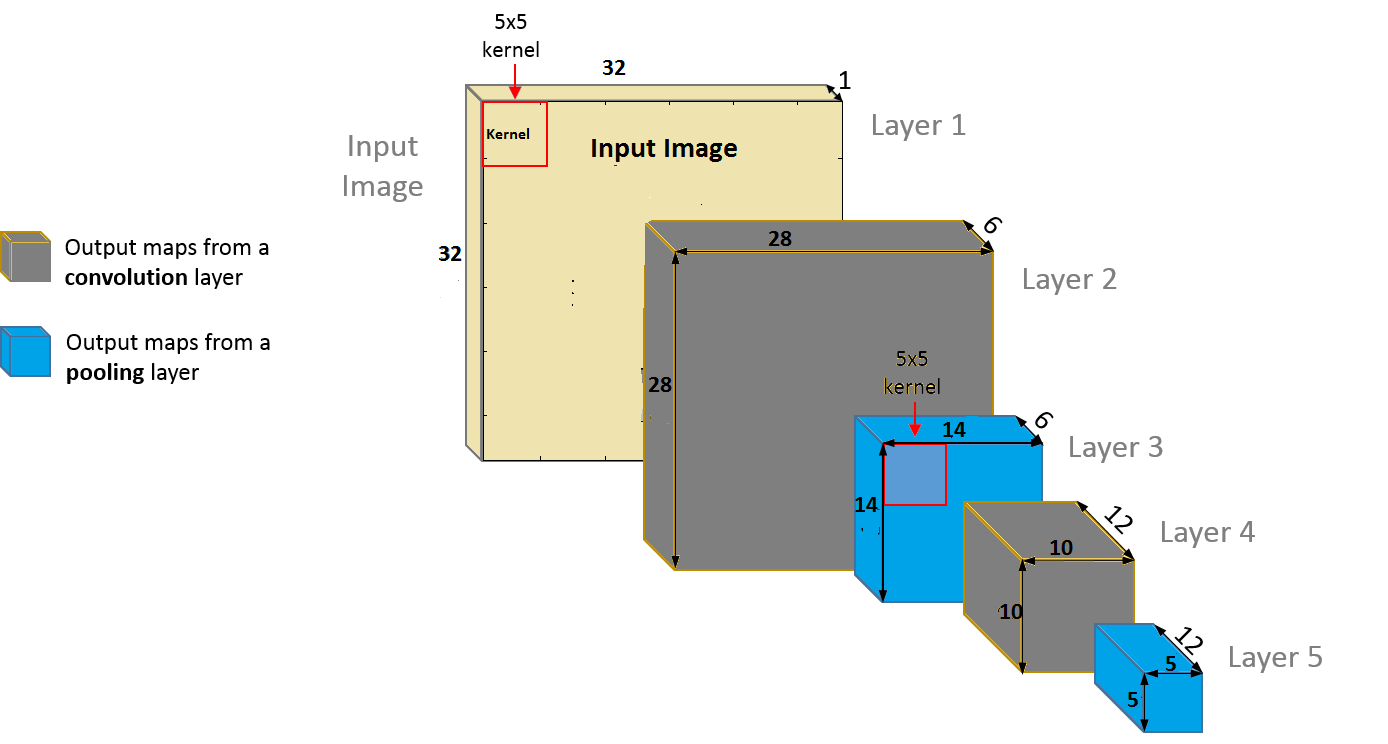}}
			
			\caption{Deep Learning architecture}
			\label{fig:DL}
		\end{figure}
\begin{enumerate}
\item \textbf{Layer 1}

		We used $4$ hidden layer deep architecture based on rasmusberg implementation to detect the fence joints. The input to the CNN  Layer $1$, is a $32\times 32$ pixel gray-scale joint images.
		
		\item \textbf{Layer 2}
		
		The first hidden layer of the network \textquotedblleft Layer 2", is going to perform some convolutions over the image using a filter mask of $5\times5$ pixels in size to yield a $28\times28$ filtered image. This layer has $6$ distinct filters that were convolved with the input image. These six convolutions will generate $6$ separate output maps, giving us a $28\times28\times6$ matrix as the output of layer $2$.
		
		\item	\textbf{Layer 3}
		
			Layer $3$ is a pooling layer which has the effect of sub-sampling the output maps by a factor of $2$ in both dimensions, so we get a $14\times14\times6$ matrix.
			
		\item	\textbf{Layer 4}
		
			Layer $4$ is another convolution layer again with a kernel size of $5\times5$ of $12$ distinct filters. To apply a single Layer $4$ filter, we actually perform $6$ convolutions (one for each output map in Layer $3$), and then sum up all of the resulting maps to make a single $10\times10\times1$ output map for that filter. This is done for each of the $12$ filters to create the $10\times10\times12$ output of Layer $4$. 
			
		\item	\textbf{Layer 5}
		
			Finally, we perform one last pooling operation that's identical to the one in Layer 3.The resulting output maps are unwound into our final feature vector containing $300$ values which is then classified using simple linear classifiers on the output. The architecture is shown in Fig.~\ref{fig:DL} with convolutional and pooling layers.
		
\end{enumerate}

\subsection{CNN training}

\textbf{Data augmentation} is a method applicable to shallow and deep representations, but that has been so far mostly applied to the latter. By augmentation it means perturbing an image I by transformations that leave the underlying class unchanged (e.g. cropping and flipping) in order to generate additional examples of the class. Augmentation can be applied at training time, at test time, or both. The augmented samples can either be taken as-is or combined to form a single feature, e.g. using sum/max-pooling or stacking. The images are downsized so that the smallest dimension is equal to $27$ pixels and a $27\times27$ crop is extracted from the center. The other strategy is to use flip augmentation, mirroring images about the y-axis producing two samples from each image. The final training of CNN is done using $20,000$ positive images and $40,000$ negative images with \emph{batchsize of $50$ and  $100$ epochs}.

A convolutional neural network consists of typically 3 different types of layers.

		 \textbf{Convolutional Layers}:
		 
		  Convolutional layers consist of a rectangular grid of neurons. It requires that the previous layer also be a rectangular grid of neurons. Each neuron takes inputs from a rectangular section of the previous layer; the weights for this rectangular section are the same for each neuron in the convolutional layer. Thus, the convolutional layer is just an image convolution of the previous layer, where the weights specify the convolution filter.
		In addition, there may be several grids in each convolutional layer; each grid takes inputs from all the grids in the previous layer, using potentially different filters.
		
	\textbf{Max-Pooling}:
	
	After each convolutional layer, there may be a pooling layer. The pooling layer takes small rectangular blocks from the convolutional layer and subsamples it to produce a single output from that block. There are several ways to do this pooling, such as taking the average or the maximum, or a learned linear combination of the neurons in the block. Our pooling layers will always be max-pooling layers; that is, they take the maximum of the block they are pooling.
	
		 \textbf{Fully-Connected}: 
		 
		 Finally, after several convolutional and max pooling layers, the high-level reasoning in the neural network is done via fully connected layers. A fully connected layer takes all neurons in the previous layer (be it fully connected, pooling, or convolutional) and connects it to every single neuron it has. Fully connected layers are not spatially located anymore (you can visualize them as one-dimensional), so there can be no convolutional layers after a fully connected layer.

The training of CNN is usually done through conventional forward and backward propagation. We briefly analyze the algorithms to do prediction and gradient computations in these architecture.

 \textbf{1. Forward Propagation}

The forward and backward propagations will differ depending on the  layer we are propagating through. The training of samples in various layers is briefly explained below.

\textbf{Convolutional Layers}

 Suppose that we have some $N\times N$ square neuron layer which is followed by our convolutional layer. If we use an $m\times m$ filter $\omega$, our convolutional layer output will be of size ${(N-m+1)}\times{(N-m+1)}$. In order to compute the pre-nonlinearity input to some unit ${x}_{ij}$ in our layer, we need to sum up the contributions (weighted by the filter components) from the previous layer cells:

\textbf{Max-Pooling Layers}

The max-pooling layers are quite simple, and do no learning themselves. They simply take some $k\times k$ region and output a single value, which is the maximum in that region. For instance, if their input layer is a $N\times N$ layer, they will then output a $N/k\times {N/k}$ layer, as each $k\times k$ block is reduced to just a single value via the max function.

\textbf{2. Backward propagation}

 The error function, E, is the mean sum of residual square errors and we have the error values at our convolutional layer. We need to find the  error values at the layer before it, and the gradient for each weight in the convolutional layer. The back-propagation algorithm in a normal fully connected layer is briefly explained below
 
 \begin{enumerate}
 
\item 1. Compute errors at the output layer L:
        
       \[ \frac{\partial E}{\partial y_i^L}
       =  \frac{\partial E(y^L)}{\partial y_i^L}
        \]

\item  2. Compute partial derivative of error with respect to neuron input (deltas) at first layer ℓ that has known errors:
  \[ \frac{\partial E}{\partial x_j^L}
          = \sigma (x_j^L) \frac{\partial E}{\partial y_j^L}
         \] 

\item 3. Compute errors at the previous layer (backpropagate errors):
 \[ \frac{\partial E}{\partial y_i^{l-1}}
 =   \sum w_{ij}^l  \frac{\partial E}{\partial x_j^{l}}
 \] 
\item 4. Repeat steps 2 and 3 until deltas are known at all but the input layer.

\item 5. Compute the gradient of the error (derivative with respect to weights):
 \[ \frac{\partial E}{\partial w_{ij}^L}
 =  y_i^{l}  \frac{\partial E}{\partial x_j^{l+1}}
 \] 

\item 6. Update the weights of filter using learning rate  
  \[  w_{ij}^L
  =     w_{ij}^L  - \alpha  \frac{\partial E}{\partial x_j^{l+1}}
  \]

Note that in order to compute derivatives with respect to weights in a given layer, we use the activations in that layer and the deltas for the next layer. Thus, we never need to compute deltas for the input layer.

\end{enumerate}

	\begin{table}[!htb]

		\centering
		\begin{tabular}{c c c c }
			\includegraphics[width=3.5cm,height =3cm]{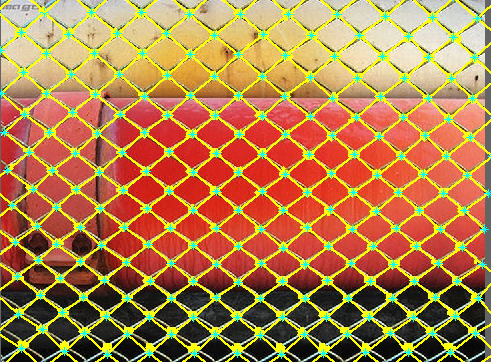} &
			\includegraphics[width=3.5cm,height =3cm]{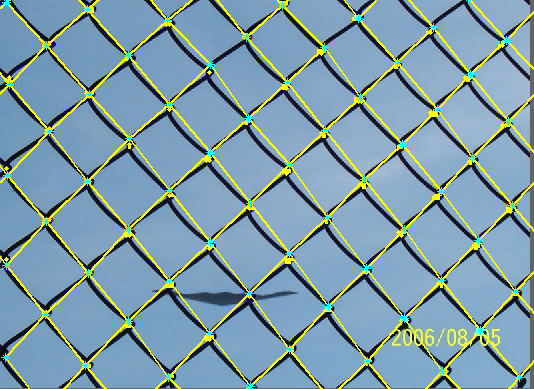} &
			\includegraphics[width=3.5cm,height =3cm]{f3.png} &
			\includegraphics[width=3.5cm,height =3cm]{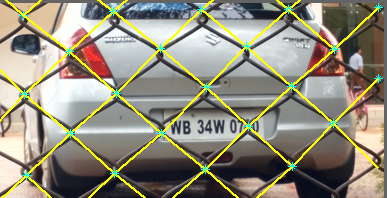} \\

			\includegraphics[width=3.5cm,height =3cm]{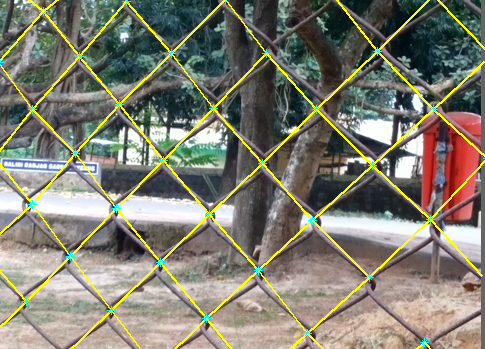} &
			\includegraphics[width=3.5cm,height =3cm]{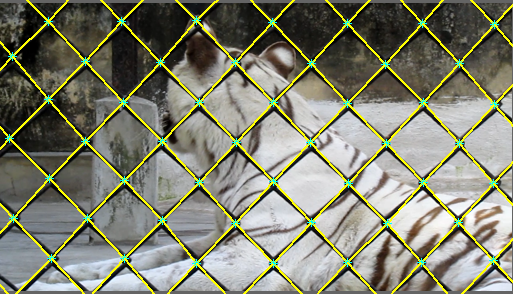} &
			\includegraphics[width=3.5cm,height =3cm]{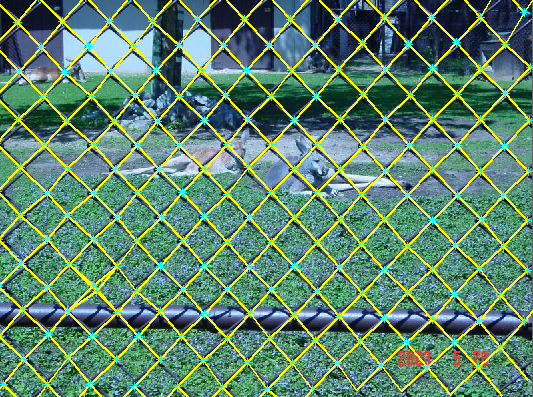} &
			\includegraphics[width=3.5cm,height =3cm]{f10.png} 		\\
		
			\includegraphics[width=3.5cm,height =3cm]{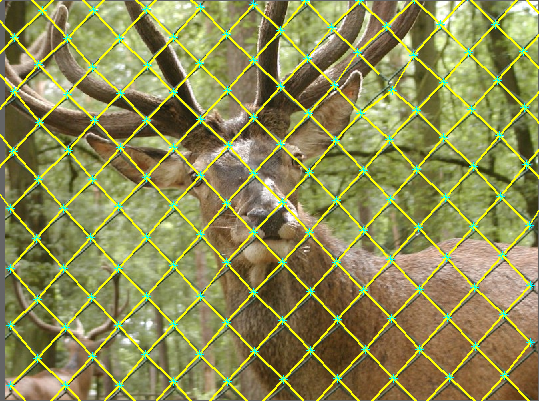} &
			\includegraphics[width=3.5cm,height =3cm]{f66.png} &
			\includegraphics[width=3.5cm,height =3cm]{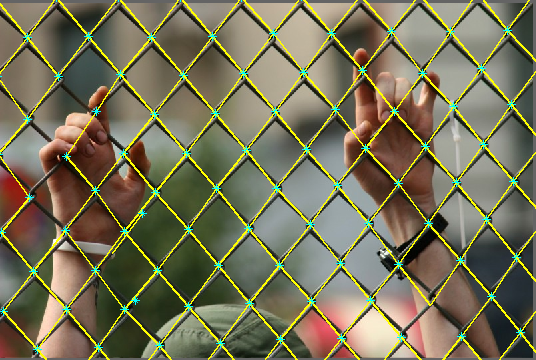} &
			\includegraphics[width=3.5cm,height =3cm]{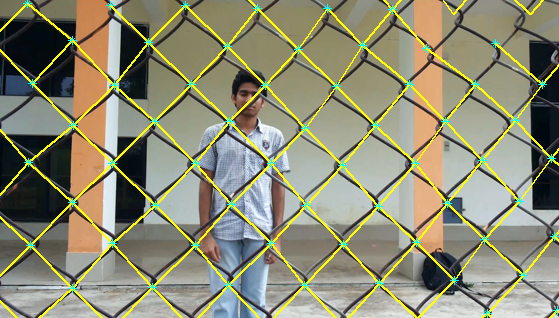} \\

		\end{tabular}
		\caption{Detected fences using deep learning approach}
		
	\end{table}

\chapter{Motion estimation}

We assume that there exists relative motion between the scene and the camera either because the user pans his camera while recording the video so as to cover the entire scene that is occluded by the fence or when the objects in the scene move with respect to a static camera. We can therefore hope that the part of the scene that is hidden behind the fence in one frame will become visible in another frame. We need to estimate the relative motion of pixels between frames being used for filling in missing information in the reference frame. 

The motion is  estimated by assuming it to be Affine model. The affine motion accounts for translation, rotation and zooming effects in the Image.The affine SIFT algorithm \cite{Guoshen} is implemented on the two frames of the video and the corresponding Interest Points are found. We use the technique of least square Fit  to calculation affine Motion matrix. 

\begin{figure}[h]
	\centering
 {\includegraphics[width=10cm]{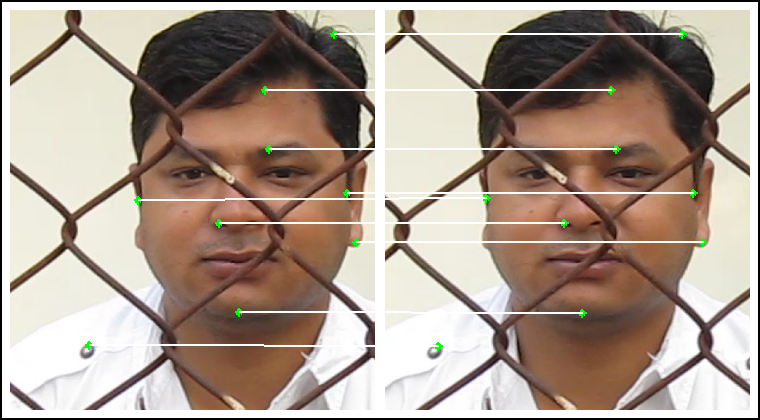}}

	\caption{SIFT feature point matching between two frames}
		\label{fig:sift}
\end{figure}

\begin{figure}[h]
	\centering
	{\includegraphics[height=4cm,width=14cm]{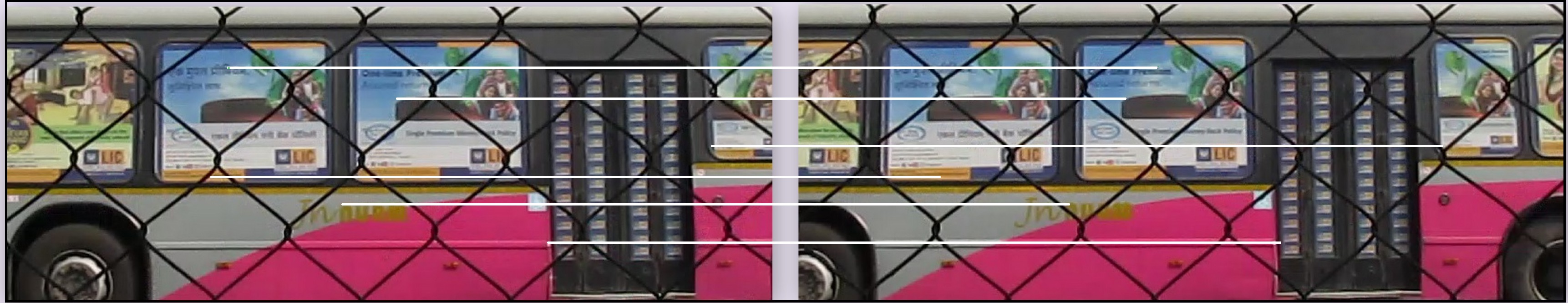}}

	\caption{SIFT feature point matching between two frames}
	\label{fig:sift}
\end{figure}

The Affine Motion model $Y= AX +B$ is given by Eq. \ref{eq:Affine model}. The motion between the points $P1(x_1,x_2)$, $P2(y_1,y_2)$ in two frames is calculated by using affine motion matrix.

  A=
  $\begin{pmatrix}
  	a_{11}       & a_{12} \\
  	a_{21}       & a_{22} \\
  	
  \end{pmatrix}$
  , accounts for rotation effects in the image and
  
 B= 
 $\begin{pmatrix}
 b_{11} \\
 b_{12} 
 \end{pmatrix}$
 , accounts for translation effects in the image

\begin{equation}
\begin{pmatrix}
y_{1}      \\
y_{2}
\end{pmatrix}
=
\begin{pmatrix}
a_{11}       & a_{12} \\
a_{21}       & a_{22} \\
\end{pmatrix}  \times \begin{pmatrix}
x_{1}    \\
x_{2}   
\end{pmatrix}
+\begin{pmatrix}
b_{11}    \\
  b_{12} 
\end{pmatrix}
\label{eq:Affine model}
\end{equation}

The Feature points are then matched in two frames using semi-automatic approaches ad the corresponding pixel is estimated for each object in the video.

\section{Large displacement optical flow}

Now, we have explored advanced motion estimated techniques between frames using Large displacement optical by \cite{Brox}. In real world situations, the shifts are non-global in nature. To apply our approach for the dynamic scenes we need the sub-pixel motion between the
frames. The large displacement optical flow uses the concept of sub-pixel shift between the frames to estimate the motion. The code implements a coarse-to-fine variational frame work for optical flow estimation that incorporates descriptor matches in addition to
brightness and gradient constancy constraints. These descriptor matches help flow estimation under large displacements of small structures, missed in the coarse-to-fine pyramid. Descriptor matches were obtained by matching densely sampled HOG descriptors in the two images via approximate nearest neighbor search. We also processed the image by median filter smoothing at the fenced pixels so that estimated wont be arbitrarily
discontinuous.

Brox et al. \cite{Brox} proposed an optical flow estimation, where they have integrated descriptor matching in a variational framework given in Eq. \ref{eq:a} First term in optimization framework corresponds to color and gradient consistency,second term is the smoothness constraint which is the total variation flow field and third term is the descriptor matching. This method is very effective in detecting sub-pixel motion in real world scenarios without occlusions. However for our application we need to accurately estimate the optical flow between the images with occlusions.When the optical flow for such images are estimated, we observe the erroneous values around the fenced pixels. To avoid these errors, we smoothen observations using a Gaussian kernel, prior to estimate optical flow.

\begin{equation}
%\begin{align*}
\begin{split}
E(\textbf{w}) = \int_\Omega  \Psi (\mid \textbf{I}_{2}(\textbf{x}+\textbf{w}(\textbf{x}))-\textbf{I}_{2}(\textbf{x}) \mid^{2} 
+ \gamma \mid  \nabla \textbf{I}_{2}(\textbf{x}+\textbf{w}(\textbf{x}))  - \nabla \textbf{I}_{1}(\textbf{x}+\textbf{w}(\textbf{x})) \mid^{2})d\textbf{x} \\
+  \alpha  \int_\Omega \Psi(\mid \nabla u(\textbf{x}) \mid^{2}  +  \mid \nabla v(\textbf{x}) \mid^{2})d\textbf{x} 
+   \beta \int_\Omega \delta (\textbf{x})  \rho (\textbf{x}) \Psi (\mid \textbf{w}(\textbf{x})  - \textbf{w}_{descr}(\textbf{x}) \mid^{2})d\textbf{x}
\end{split}
%\end{align*}
\label{eq:a}
\end{equation}

\begin{figure}[H]
	
	\centering
	
	\subfigure[]{\includegraphics[height=4cm,width=4cm]{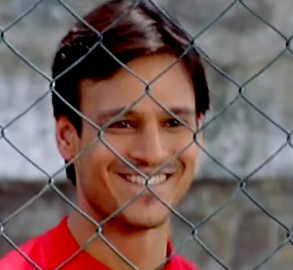}} \hspace{0.5cm}
	\subfigure[]{\includegraphics[height=4cm,width=4cm]{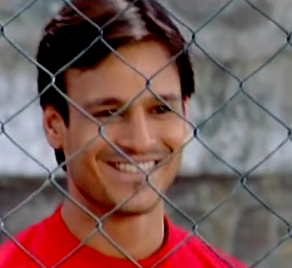}} \hspace{0.5cm}
	\subfigure[]{\includegraphics[height=4cm,width=4cm]{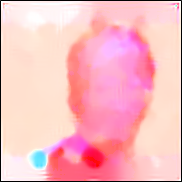}}\hspace{0.5cm}

	\caption{ Motion estimation:(a),(b) are fenced frames
		(c) flow map between two frames}
	\label{fig:flowmap}
\end{figure}

\chapter{Optimization framework}

A naive approach is to simply inpaint the individual frames of the input video by a standard image completion technique. However, such an approach would approximate missing information by propagating neighboring pixel intensities taking care of  edges/discontinuities in every frame.      	

Given $n$ frames and the corresponding fence masks, the loopy belief technique iteratively outputs the de-fenced image pixels to be filled-in are known and marked by the user in the input image. 

We model the \textquotedblleft de-fenced” image as a Markov random field and use the loopy belief propagation technique \cite{Felzenszwalb} to optimize an appropriately formulated objective function. Our approach is more accurate than mere image inpainting since we use data from neighboring frames to derive the maximum a-posteriori estimate of the  \textquotedblleft de-fenced” image. Although, there have been previous attempts for removing fences from images \cite{Park}, the novelty in our approach is formulating an optimization-based framework to fuse data from multiple frames of the input video in order to fill-in fence pixels.

We modeled the image occluded by fences as 
\begin{equation}
\textbf{y}_{m} = \textbf{O}_{m}\textbf{W}_{m}\textbf{x}+\textbf{n}_{m}
\end{equation}
where  $\textbf{y}_{m}$ represents the $m^{\mbox{th}}$ frame of the video,
$\textbf{O}_{m}$ is the binary fence mask corresponding to $m^{\mbox{th}}$ frame,
$\textbf{x}$ is the de-fenced image,
$\textbf{W}_{m}$ is the warp matrix and
$\textbf{n}_{m}$ is the noise assumed as Gaussian.

In computer vision and image processing contextual constraints such as pixel information and image features are well used to understand an image. Such constraints are modeled by well known graphical Markov random fields. We propose to model de-fenced image as a Markov random field and also we formulate optimization framework for the same. The maximum a-posteriori estimate of the de-fenced image can be obtained as

\begin{equation}
\hat{\textbf{x}}=\arg\min_{\textbf{x}} {\|\textbf{y}_{m}-\textbf{O}_{m}\textbf{W}_{m}\textbf{x}\|^2} + {\lambda{\sum_{{c}\in{C}}\textbf{V}_c(\textbf{x})}}
\label{eq:loopy}
\end{equation}
\begin{equation}
P(\textbf{x}) = \frac{1}{Z}exp(-\textbf{V}_{c}(\textbf{x}))
\end{equation}

where $\textbf{V}_{c}( \textbf{x}) = |x_{i,j} - x_{i,j +1}| + |x_{i,j} - x_{i,j - 1}| +|x_{i,j} - x_{i - 1,j}| +|x_{i,j} - x_{i +1,j}|$ is clique potential function assuming a first order neighborhood. We minimize the objective function in Eq. \ref{eq:loopy} by using the LBP technique. The parameter $\lambda$ is chosen as $5\times 10^5 $ for all our experiments.

\section{Loopy belief propagation algorithm} 

MRF are graphical model that encode spatial dependency. The problem of de-fencing may be modeled using an MRF.  It calculates the marginal distribution for each unobserved node, conditional on any observed nodes. Every pixel in the image is considered as a random variable.  Different random variable together forms a grid of random variables. Markov Random field give rise to good, flexible, stochastic image models. Over the past few years there have been exciting advances in the development of algorithms for solving early vision problems such as stereo, optical ﬂow and image restoration using MRF models. We implement the technique of loopy belief for the problem of  inpainting.

Belief propagation is a message passing algorithm. It is used to compute marginals of the latent nodes of underlying graphical model. A node passes a message to an adjacent node only when it has received all incoming messages, excluding the message from the destination node to itself.  Below Fig.~\ref{fig:messagepassing} shows an example of a message being passed from one node to other:       	
\begin{figure}[h]
	\centering
	{\includegraphics[width=5cm]{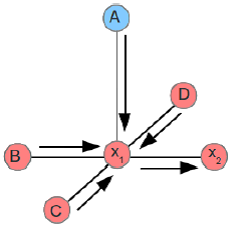}}
	\caption{Message passing to a node from neighboring nodes}
		
		\label{fig:messagepassing}
\end{figure}

The general framework for the problems we consider can be defined as follows. Let P be the set of pixels in an image and L be a set of labels. The labels correspond to quantities that we want to estimate at each pixel, such as disparities, intensities or ﬂow vectors. A labeling f assigns a label $fp\in{L}$ to each pixel $p\in{P}$.

 We assume that the labels should vary smoothly almost everywhere but may change dramatically at some places such as object boundaries. The quality of a labeling is given by an energy function,

\begin{equation}
{ \textbf{E}(f)= {\sum_{(p,q)\in{N}} \textbf{V}(f_p ,f_q)} + {\sum_{p\in{P}}\textbf{D}_p(f_p)}}
\label{eq:energyloop}
\end{equation}

where N are the edges in the four-connected image grid graph. $\textbf{V}(f_p ,f_q)$ is the cost of assigning labels $f_p$ and $f_q$ to two neighboring pixels, and is normally referred to as the discontinuity cost. $\textbf{D}_p(f_p)$ is the cost of assigning label fp to pixel p, which is referred to as the data cost. Finding a labeling with minimum energy corresponds to the MAP estimation problem for an appropriately defined MRF.

The BP algorithm works by passing messages around the graph defined by the four-connected image grid. Each message is a vector of dimension given by the number of possible labels. Let $msg_{pq}^t$ be the message that node p sends to a neighboring node q at time t. When using negative log probabilities all entries in $m_{pq}$ are initialized to zero, and with at each iteration new messages are computed in the following way:

\clearpage
\begin{enumerate}

\item \textbf{Message passing}	

\begin{equation}
msq_{pq}^t (f_q)  =   {\min_{f_p \in {L}} ( \textbf{S}(f_p ,f_q)+ \textbf{D}(f_q)  +  {\sum_{o\in{N}(p)-q)} msg_{op}^{t-1}(f_p)} ) }
\label{eq:message}
\end{equation}

\item \textbf{Belief calculation}:

After T iterations a belief vector is computed for each node,
\begin{equation}
belief_q(f_q)= D_q(f_q) +  \sum_{p\in N(q))}msg_{pq}^t (f_q)
\label{eq:belief}
\end{equation}

\end{enumerate}

\begin{algorithm}[h]
	\caption{Image de-fencing algorithm}\label{alg:de-fencing}
	\begin{algorithmic}[1]
		\State $\textbf{Input:}\lambda, L,\textbf{O}_{m},\textbf{Y}_{m}$
		\State $Initialise: All\hspace{5pt} messages\hspace{5pt} to\hspace{5pt} zeros$
		\State $D(f_{p})=\sum_{m}(\textbf{O}_{m}.*(f_{p}-\textbf{Y}_{m})^2)$
		\While{\(t\leq T\)}
		\For{\(f_{q} = L\)}
		\State $ S= \lambda \mid L-l_{q} \mid$
		\State $ msg_{pq}^t(f_{q})= \min(D+S+\sum_{o\in N(p)-q}msg_{op}^{t-1}(f_{p}))$
		\State $ t\gets t+1$
		\EndFor
		\EndWhile
		\For{\(f_{q} = L\)}
		\State $belief_{q}(f_{q})=D_{q}(f_{q})+\sum_{p\in N(q)}msg_{pq}^T(f_{q})$
		\EndFor
	\end{algorithmic}
\end{algorithm}

\vspace*{1cm}
Finally, the label $f_q$ that minimizes ${belief}_q(f_q)$ individually at each node is selected. The standard implementation of this message passing algorithm on the grid graph runs in $O(nk^{2}T)$ time, where n is the number of pixels in the image, k is the number of possible labels for each pixel and T is the number of iterations. Basically it takes $O(k^{2})$ time to compute each message and there are O(n) messages per iteration. 	For the current problem, the number of labels is number of intensities available i.e $255$.  The Data Cost term returns the cost/penalty of assigning any label value to a data. The Smoothness Cost function, enforces smooth labeling across adjacent hidden nodes and to do this it penalizes adjacent labels that are different. 

\chapter{Results}

In this section, we first evaluate the performance of our proposed lattice detection algorithm on two different datasets. Next we discuss the results obtained using our optimization framework on synthetic as well as real world data. To validate our approach, we compare with the state-of-the art inpaintings as well as de-fencing techniques. We present results of experiments using several real-world videos to demonstrate the effectiveness of the proposed algorithm.

\section{Fence Detection}

\subsection{Dataset}	
 Unfortunately, there is no benchmark fence data set available in the literature for evaluating the proposed learning based algorithm. For demonstrating the effectiveness of the proposed learning based algorithm we used two different data sets. Firstly, we have collected a dataset consisting of $200$ real-world images/videos under diverse scenarios and complex backgrounds by using a mobile camera (Google Nexus 5). Secondly, we used a subset of images from Penn State University (PSU) near-regular texture (NRT) database \cite{Minwoo}.
 
  The images in NRT database are divided into three categories wherein data set $1$ (D1) contains $67$ images with opaque texels and appearance variations of the repeating elements due to different viewpoint and lighting conditions. Data set $2$ of the NRT database (D2) contains $73$ images with see-through or wiry structures. Data set $3$ (D3) contains $121$ images with views of city buildings having multiple repeating patterns with perspective distortion. However, only a subset of $40$ images from D2 are of fences. We report results of the proposed machine learning approach for detection of fences on these $40$ images of D2 dataset in the NRT database. The comparative results of fence detection is shown in Table~\ref{tab:PerformanceEval1}. The Deep learning based fence detection approach works with a precision of 0.97 on NRT database and 0.88 on Our database
  
  \subsection{Comparison with state-of-the-art}
  The Precision measure is defined as ratio the detected true  positives(tp) to the ground truth positives. The Recall is defined as the ratio of detected false positives(fp) to the ground truth positives. 
  The traditional F-measure or balanced F-score (F1 score) is the harmonic mean of precision and recall.

 \begin{equation}
 \text{Precision}=\frac{tp}{tp+fp}  ,     Recall=\frac{tp}{tp+fn} 
 \end{equation}

\begin{equation}
F-measure = 2 \cdot \frac{\mathrm{precision} \cdot \mathrm{recall}}{\mathrm{precision} + \mathrm{recall}}
\end{equation}

\begin{table}[h]
	\caption{Quantitative evaluation of fence detection}% title name of the table
	\vspace*{2mm}
	\centering
	%\renewcommand{\arraystretch}{2.0} \setlength{\tabcolsep}{1.8pt} \footnotesize
	% centering table
	\scalebox{0.85}
	{
		
		\begin{tabular}{|c|c|c|c|c|c|c|}
			% creating 10 columns
			\hline
			% inserting double-line
			& \multicolumn{3}{|c|}{Our Database} & \multicolumn{3}{|c|}{NRT Database }\\\cline{2-7}
			Method & Precision & Recall & F-measure & Precision & Recall & F-measure\\
			\hline
			Park \cite{Minwoo} & 0.94 & 0.26 & 0.41 & 0.95 & 0.46 & 0.62\\
			\hline
			
			\textbf{HOG+SVM method} & 0.96 & 0.95 & \textbf{0.96} & 0.95 & 0.92 & \textbf{0.93}\\
			\hline
			
			\textbf{Deep learning} & 0.88 & 0.964 & 0.92 & 0.97 & 0.96 & \textbf{0.965}\\
			\hline
		\end{tabular}
		
	}
		\label{tab:PerformanceEval1}
	\end{table}

\section{Image de-fencing}
Initially, we report the results on synthetic data where the fence positions and object movement are already known. Later we demonstrate the effectiveness of our algorithm on real world scenarios. 

\subsection{Synthetic data}
\begin{figure}[h]
	
	\centering
	
	\subfigure[]{\includegraphics[width=6cm]{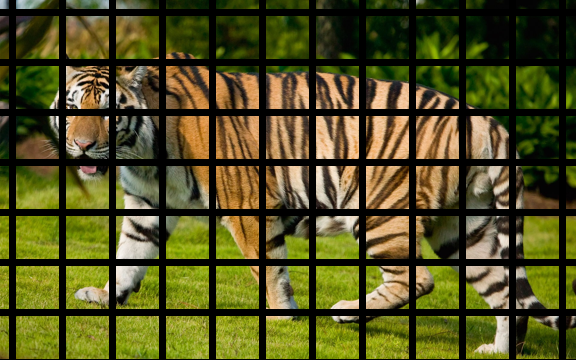}} \hspace{0.5cm}
	\subfigure[]{\includegraphics[width=6cm]{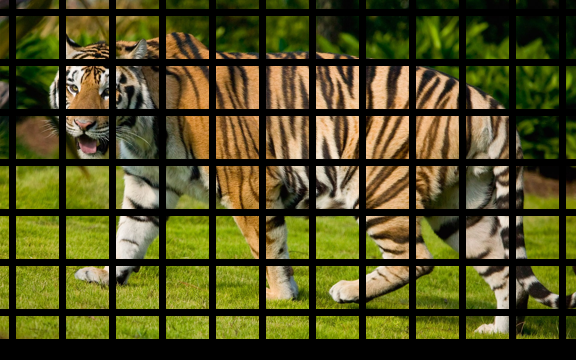}}\\
	\subfigure[]{\includegraphics[width=6cm]{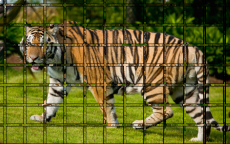}}
	\hspace{0.5cm}
	\subfigure[]{\includegraphics[width=6cm]{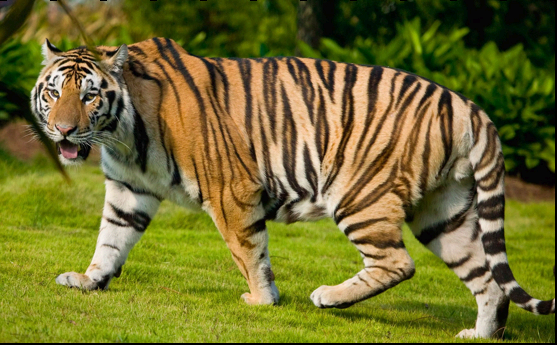}}

	\caption{ Image de-fencing Synthetic case: (a),(b) are fenced frames
		 (c) Output by \cite{Criminisi} method (d) Proposed method output }
		\label{fig:syntiger}
\end{figure}

Initially, we report the results on synthetic data. Here we assumed that the locations of fence pixels were known and that the fence was static. Only the background is shifted with respect to the static fence. The fenced observations on the synthetic data is shown in Fig.~\ref{fig:syntiger} (a),(b). Note that there are hardly any artifacts and the fence has been successfully filled in. The Root Mean Square error (RMSE) is $3.17$, Peak Signal to noise ratio (PSNR) is $38.11$ dB and the Structural Similarity Index (SSIM) is $0.87$. The quantitative comparative results of our algorithm with the state-of-the-art inpainting techniques are shown in Table~\ref{tab: synresults}. The de-fencing results on a synthetic data is shown in Fig.~\ref{fig:syntiger} (d). 

\begin{table}[!htb]
	\caption{Quantitative evaluation of inpainting \& de-fencing}
	\begin{center}
		\begin{tabular}{|l|l|l|l|}
			\hline
			Algorithm  & PSNR & RMSE & SSIM\\
			\hline
			Exemplar-based inpainting\cite{Criminisi} & 21.67 & 4.70 & 0.85 \\
			\hline
			Total variation inpainting\cite{Getreuer} & 26.06 & 3.75 & 0.93 \\
			\hline     
			\textbf{Our method} & \textbf{49.50} & \textbf{0.40} & \textbf{0.99}\\
			\hline
		\end{tabular}
		
		\label{tab: synresults}
	\end{center}
\end{table}

\clearpage

\subsection{Real-world scenarios}

\begin{figure}[H]
		
		\centering
		
		\subfigure[]{\includegraphics[height=4.5cm,width=4cm]{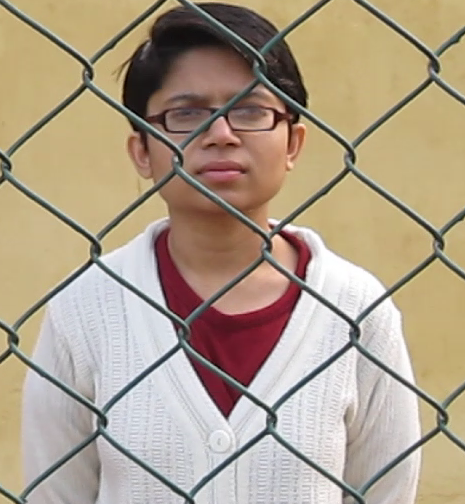}}\hspace{1cm}
		\subfigure[]{\includegraphics[height=4.5cm,width=4cm]{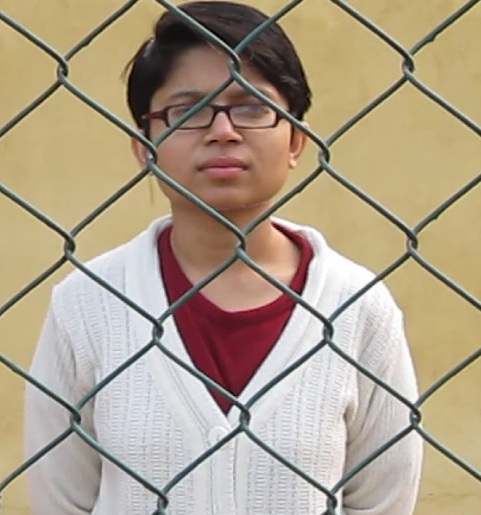}}\\
		\subfigure[]{\includegraphics[height=4.5cm,width=4cm]{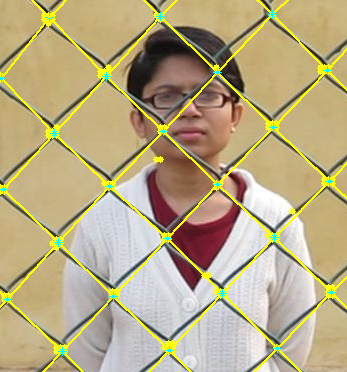}}\hspace{1cm}
		\subfigure[]{\includegraphics[height=4.5cm,width=4cm]{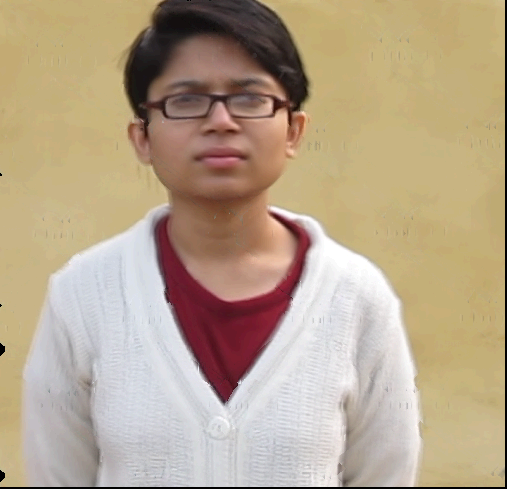}}

		\caption{ Real world scenario: Girl standing behind fence (a),(b) are fenced frames (c) Detected fence (d) De-fenced image }
		\label{fig:vrushali}
	\end{figure}

We show series of de-fencing results on real world scenarios in this section. The first experiment used a real-world video of a girl standing behind a fence by panning a camera with arbitrary smooth motion. A frame from this video is shown in Fig.~\ref{fig:vrushali} (a), (b). We used 4 frames from the captured video wherein both the fence pixels as well as the background were moving. Due to the significant depth of the scene from the panning camera, it is valid to assume that the background pixels were shifted globally by a fixed amount in the different frames.

we match the corresponding feature points  in the  first and second observations using the affine SIFT descriptor. The global shifts of the three chosen frames are (13; 33) (18; 50) (15; 60) pixels, with respect to the first observation.  We detect the fence pixels using our fence detection   technique. The  de-fenced image obtained by using the four observations in the proposed method is shown in Fig.~\ref{fig:vrushali} (d). The proposed algorithm has effectively reconstructed data even at the eyes, lips and the nose where the fence occludes the face.

\clearpage

\begin{figure}[!htb]
	
	\centering
	
	\subfigure[]{\includegraphics[width=6cm]{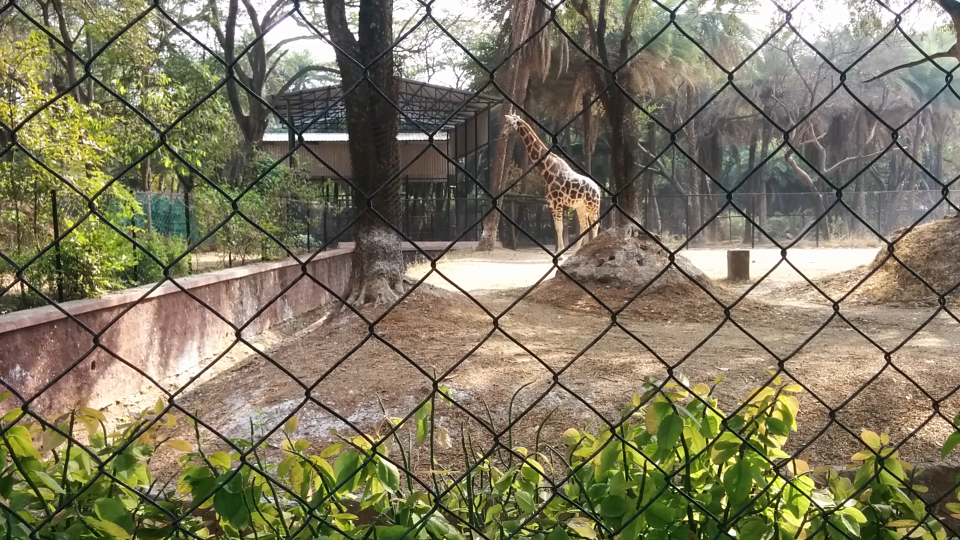}}\hspace{1cm}
	\subfigure[]{\includegraphics[width=6cm]{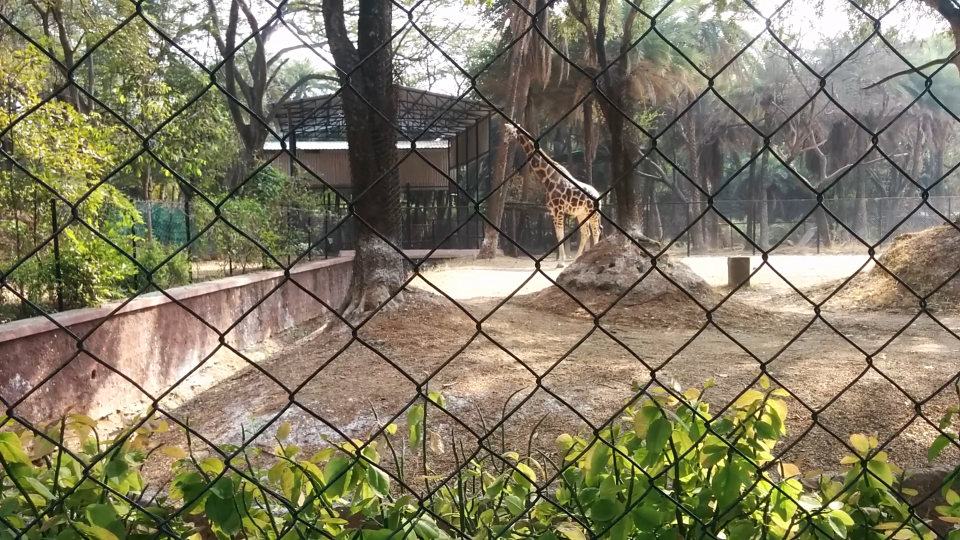}}\\
	\subfigure[]{\includegraphics[width=6cm]{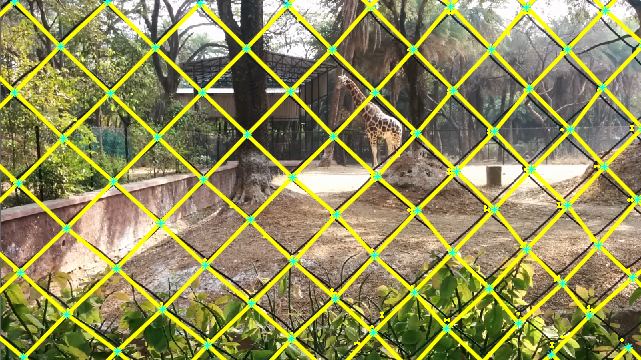}}\hspace{1cm}
	\subfigure[]{\includegraphics[width=6cm]{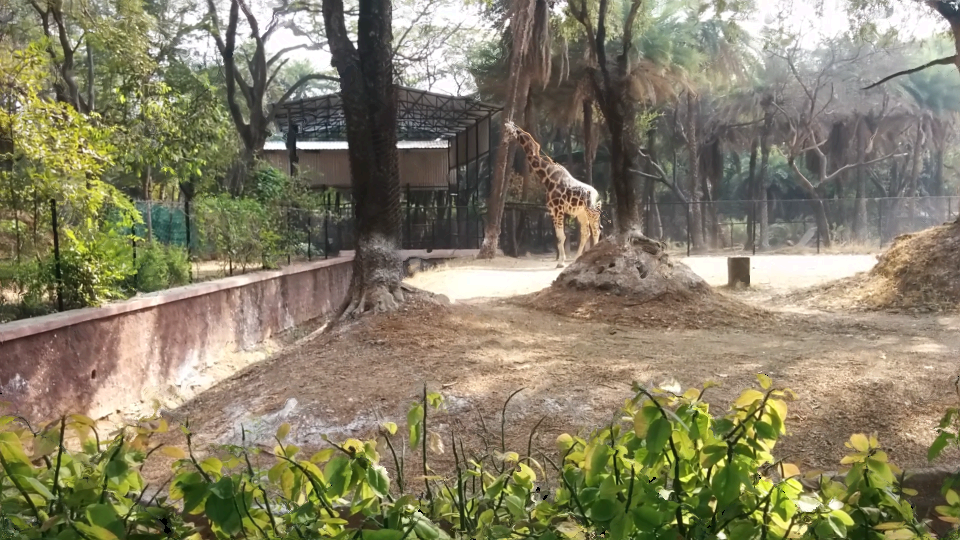}}

	\caption{ Giraffe at Hyderabad Zoo  (a),(b) are fenced frames
		(c) Original Image (d) De-fenced image }
	\label{fig:car4}
\end{figure}
	
	\begin{figure}[!htb]
		
		\centering
		
		\subfigure[]{\includegraphics[width=6cm]{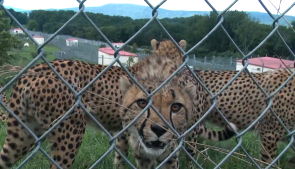}}\hspace{1cm}
		\subfigure[]{\includegraphics[width=6cm]{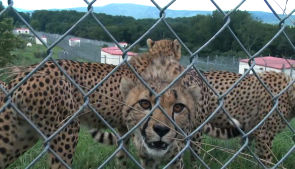}}\\
		\subfigure[]{\includegraphics[width=6cm]{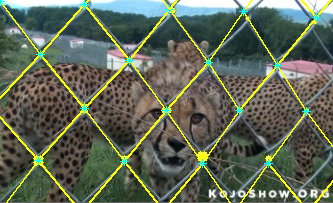}}\hspace{1cm}
		\subfigure[]{\includegraphics[width=6cm]{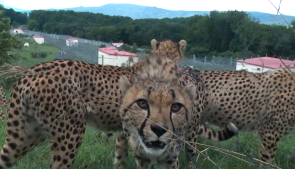}}

		\caption{ Youtube leopard data :(a),(b) are fenced frames
			(c) Detected fence (d) De-fenced image }
		\label{fig:zirafee}
	\end{figure}

In the similar way, the optimization approach is used to remove any occluded regions from Zoo video shown in  Fig.~\ref{fig:zirafee} (d)  in a image by using information from the other frames. Also, the motion estimation of every pixel can also be found by optical flow instead of affine SIFT method. We show the results of image de-fencing on various videos show in Figs. $7.2 - 7.9$. We can see that the fence occlusions are clearly reconstructed from other by using our algorithm.

\begin{figure}[!htb]
		
		\centering
		
		\subfigure[]{\includegraphics[width=6cm]{Y1.png}}\hspace{1cm}
		\subfigure[]{\includegraphics[width=6cm]{Y3.png}}\\
		\subfigure[]{\includegraphics[width=6cm]{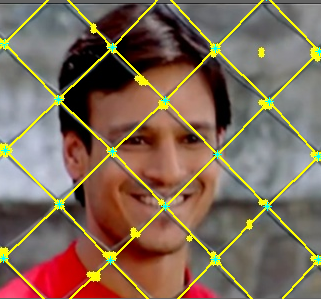}}\hspace{1cm}
		\subfigure[]{\includegraphics[width=6cm]{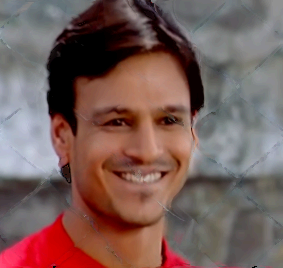}}

		\caption{ Saathiya Movie video: (a),(b) are fenced frames
			(c) Original Image (d) De-fenced image }
		\label{fig:saathiya}
	\end{figure}

\begin{figure}[!htb]
			
			\centering
			
			\subfigure[]{\includegraphics[width=6cm]{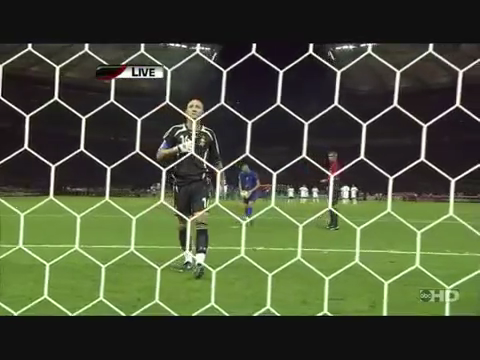}}\hspace{1cm}
			\subfigure[]{\includegraphics[width=6cm]{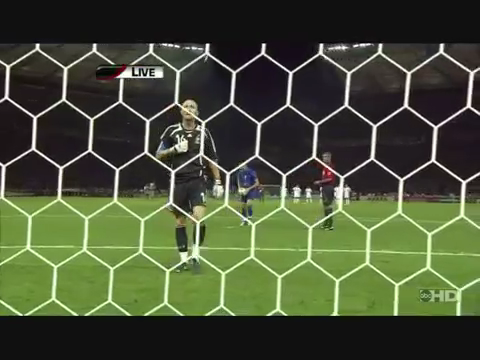}}\\
			\subfigure[]{\includegraphics[width=6cm]{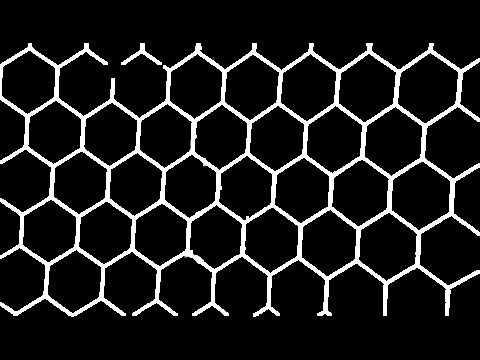}}\hspace{1cm}
			\subfigure[]{\includegraphics[width=6cm]{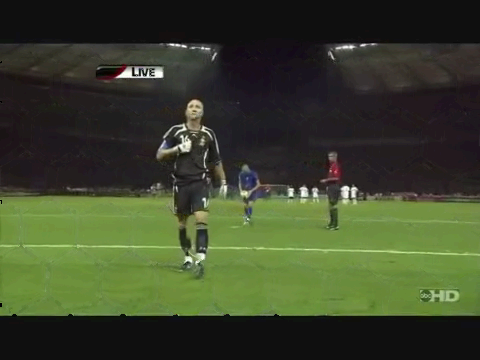}}

			\caption{ Football Video :(a),(b) are fenced frames
				(c) Detected fence (d) De-fenced image }
			\label{fig:football}
		\end{figure}

The Figs.~\ref{fig:saathiya} (a),(b) are taken from a youtube video and it can be observed that the de-fenced image Fig.~\ref{fig:saathiya} (d) is clear from all fence occlusions. In the next experiment , we work on a football video Fig.~\ref{fig:football} video and results of de-fencing is shown in Fig.~\ref{fig:football} (d).

\begin{figure}[!htb]
	
		\centering
		
		\subfigure[]{\includegraphics[width=6cm]{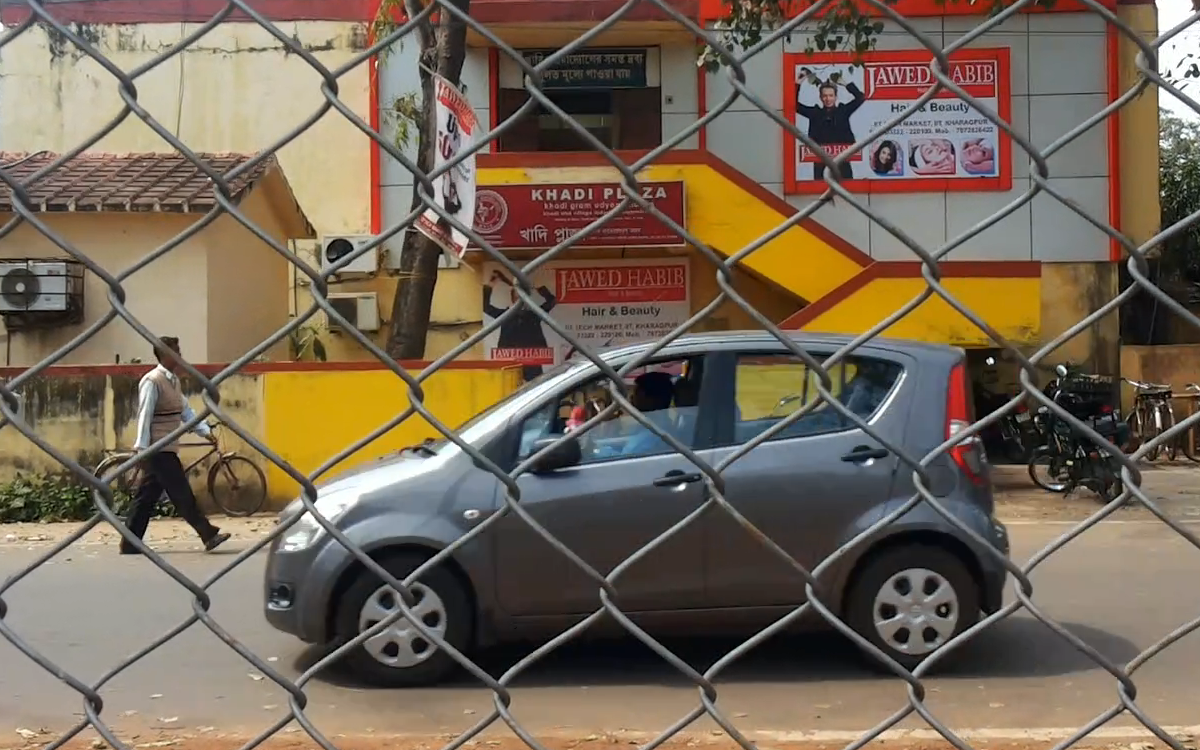}}\hspace{1cm}
		\subfigure[]{\includegraphics[width=6cm]{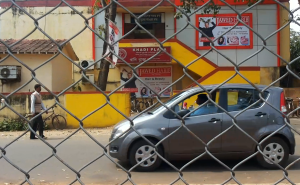}}\\
		\subfigure[]{\includegraphics[width=6cm]{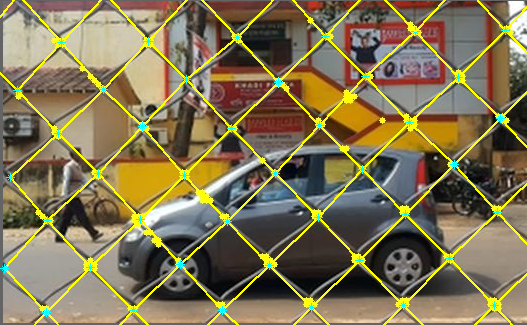}}\hspace{1cm}
		\subfigure[]{\includegraphics[width=6cm]{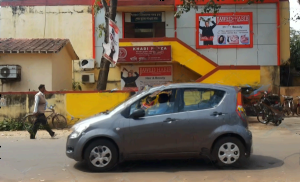}}

		\caption{ Car Video at IIT Kharagpur: (a),(b) are fenced frames
			(c) Detected fence (d) De-fenced image }
		\label{fig:kgpcar}
	\end{figure}

\begin{figure}[!htb]
		
		\centering
		
		\subfigure[]{\includegraphics[width=6cm]{car1.png}}\hspace{1cm}
		\subfigure[]{\includegraphics[width=6cm]{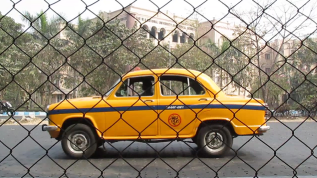}}\\
		\subfigure[]{\includegraphics[width=6cm]{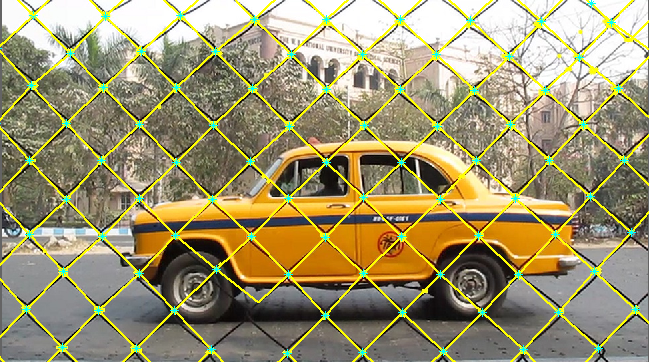}}\hspace{1cm}
		\subfigure[]{\includegraphics[width=6cm]{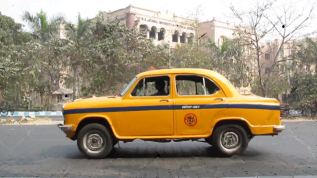}}

		\caption{ Traffic car data at kolkatta: (a),(b) are fenced frames
			(c) Detected fence (d) De-fenced image }
		\label{fig:traffickol}
	\end{figure}

\begin{figure}[!htb]
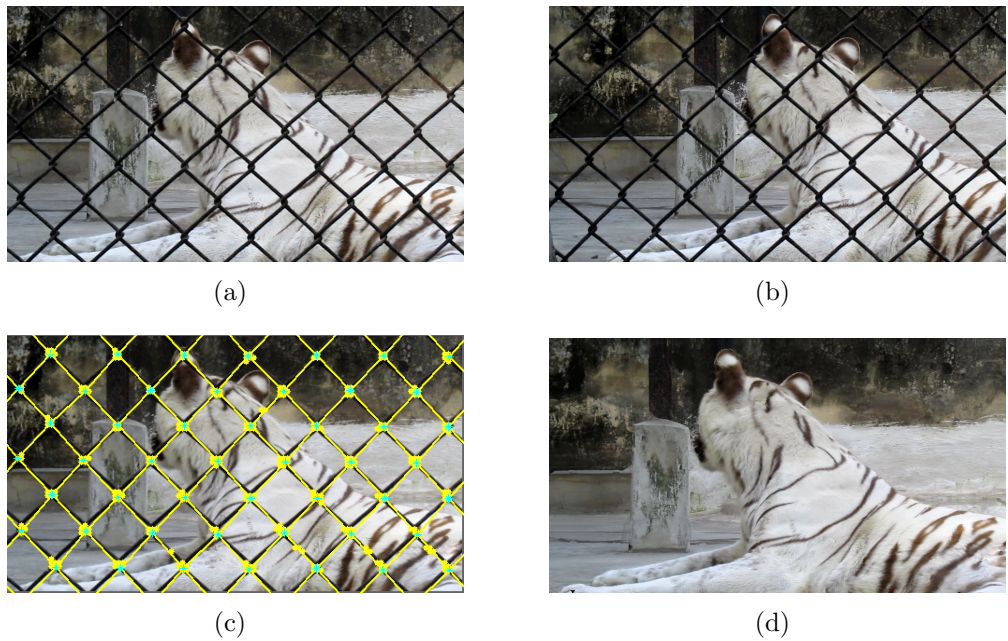

		
			\centering
			
			\subfigure[]{\includegraphics[width=6cm]{t1.png}}\hspace{1cm}
			\subfigure[]{\includegraphics[width=6cm]{t2.png}}\\
			\subfigure[]{\includegraphics[width=6cm]{tf.png}}\hspace{1cm}
			\subfigure[]{\includegraphics[width=6cm]{tr.png}}

			\caption{ Lion in Kolkatta zoo:(a),(b) are fenced frames
				(c) Detected fence (d) De-fenced image }
				\label{fig:zookol}
		\end{figure}

Similarly, we report the results on various real world scenarios such as  Traffic car videos are shown in Figs.~\ref{fig:kgpcar}, ~\ref{fig:traffickol}, and also Zoo park videos shown  in Figs. ~\ref{fig:zirafee},      ~\ref{fig:zookol}. The corresponding de-fenced images are also shown in respective figures.

\clearpage
\section{Failure cases}
	
\begin{figure}[!htb]
		
		\centering
		\subfigure[first frame]{\includegraphics[width=4 cm]{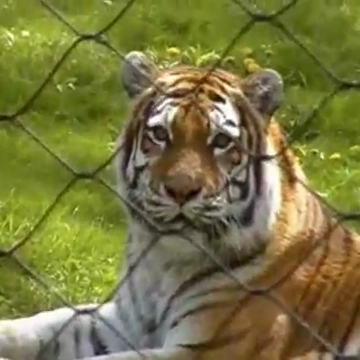}} \hspace{0.5cm}
		\subfigure[second frame]{\includegraphics[width=4 cm]{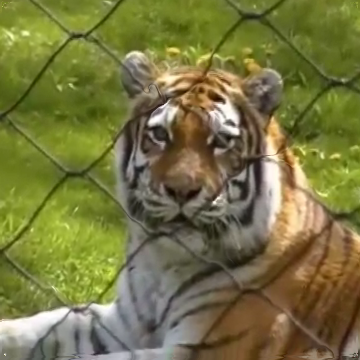}} \hspace{0.5cm}
		\subfigure[De-fenced image]{\includegraphics[width=4 cm]{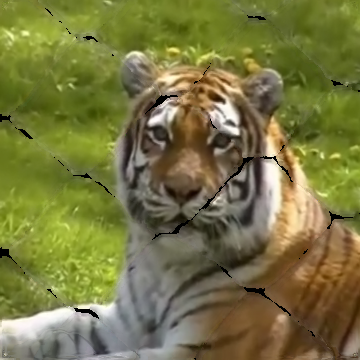}} 
		
		\caption{ Failure cases: Image de-fencing}
		\label{fig:failion}
	\end{figure}

 Our approach of de-fencing fails in some of the scenario as shown in Fig.~\ref{fig:failion}. It failed to reconstruct the fenced image from occluded videos as  there is not enough relative motion between the frames. We can observe in the Fig.~\ref{fig:failion} (c) that some of the regions have not been properly reconstructed due to no relative motion between the frames which is one of the major constraints in our algorithm. 	
 
\begin{figure}[!htb]
		
		\centering
		\subfigure[first frame]{\includegraphics[height=5cm,width=4 cm]{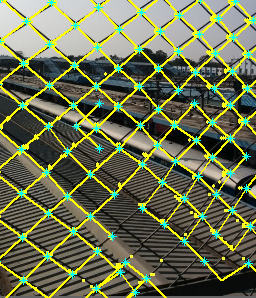}} \hspace{0.5cm}
		\subfigure[second frame]{\includegraphics[height=5cm,width=4 cm]{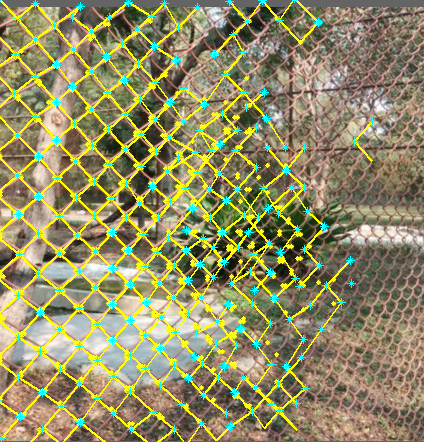}} \hspace{0.5cm}
		\subfigure[De-fenced image]{\includegraphics[height=5cm,width=4 cm]{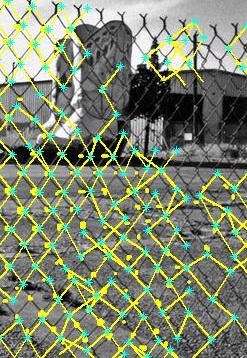}} 
		
		\caption{Failure cases: Fence detection}
		\label{fig:faildetect}
	\end{figure}

Our approach of fence detection fails in extremely tough deformed cases as the fence joint shape totally changes. Some of our failure cases is shown in Fig.~\ref{fig:faildetect}.	It was observed that CNN architecture  relatively works better than the HOG features based detection in the deformed lattice cases.

%\justify

\chapter{Conclusion}

We proposed an approach for de-fencing an image using multiple frames from a video captured by panning the scene. We divided this problem into three sub-problems: (a) detection of the fence pixels (b) estimation of motion of background pixels (c) filling up of missing data at the locations of fence pixels. We have proposed a semi-automated approaches for all the sub-problems. An optimization based framework was formulated for fusing data from multiple relatively shifted frames to fill-in missing data. Our results for both synthetic and real world data show the effectiveness of the proposed algorithm.

Automatic fence detection using machine learning approaches like SVM, CNN significantly outperforms than the available state-of-art lattice detection algorithms. We have evaluated our algorithm on an benchmark dataset NRT database and an independently developed dataset by ours. The evaluation metrics precision, recall are significantly higher than the state-of-the-art deformed lattice detection algorithm. Also, we observed the convolutional deep layer architecture has performed relatively better than HOG featured based SVM even in the tough deformed cases.

To solve the problem of motion estimation between the frames, we initially assumed that the frames of the input video obtained by panning the occluded scene are shifted globally. Hence, we used the affine scale-invariant feature transform image descriptor to match corresponding points in the frames obtained from the captured video. The above assumption restricts our algorithm to videos having global motion only and so it is not general enough for real-world data wherein scene elements could also be dynamic. In order to overcome this limitation we need an algorithm which estimates local pixel motion automatically. Therefore, we use a recently proposed optical flow \cite{Brox} technique for motion estimation in videos of dynamic scenes. The technique\cite{Brox} seem to estimate the motion in complex background scenes with fence occlusions.

The final major contribution  in our method is fusion of data from additional frames to de-fence the reference image. For this purpose, we proposed a degradation model to describe the formation of images affected by fences. We use the loopy belief propagation technique to optimize an appropriately formulated objective problem. Our approach is more accurate than mere image inpainting since we use data from neighboring frames to derive the maximum a-posteriori  estimate of the de-fenced image. We have demonstrated our de-fencing on various real world videos and compared our technique with the available inpainting techniques.

\section{Future work}

Although, the fence detection results with convolutional neural networks significantly outperforms than state-of-art lattice detection algorithms, it can still be optimized by experimenting with the architecture. The training dataset for CNN should be increased more for experimenting on deep architectures. The motion estimation using optical flow fails in some cases if the relative motion between fence and background is large. The fence detection on deformed cases can be improved by using the concept of homography by making it fronto-parallel and later applying our algorithms on it.

\section{Work publications}

\begin{enumerate}

\item J.S. Ganesh, K.K Nakka, R.R. Sahay, \textquotedblleft Towards an Automated Image De-fencing Algorithm Using Sparsity", in \textit{10th International  Conference on Computer Vision Theory and Applications}, 2015. (Accepted)

\item J.S. Ganesh, K.K Nakka, R.R. Sahay, \textquotedblleft My Camera can see through fences", in \textit{The Journal of the Optical Society of America A}, 2015. (Draft prepared)

\end{enumerate}

\cleardoublepage

\InputIfFileExists{MTP_final.bbl}

%\addcontentsline{toc}{chapter}{Bibliography}
%\bibliographystyle{ieeetr}
%\bibliography{ganesh_db}  %

\end{document}